\documentclass{article}

 \usepackage[main, final]{neurips_2026}

\usepackage[utf8]{inputenc} %
\usepackage[T1]{fontenc}    %
\usepackage{hyperref}       %
\usepackage{url}            %
\usepackage{booktabs}       %
\usepackage{amsfonts}       %
\usepackage{nicefrac}       %
\usepackage{microtype}      %
\usepackage{xcolor}         %

\usepackage{tabularx}
\usepackage{array}
\usepackage{graphicx}
\usepackage{multirow}
\usepackage{pifont}
\usepackage{amsmath}
\usepackage{wrapfig}
\usepackage{enumitem}
\usepackage{makecell}
\usepackage[ruled,vlined]{algorithm2e}
\allowdisplaybreaks

\title{FlowMoP: Stochastic Multi-Person Motion Prediction}

\author{%
  Aadya Agrawal\\ %
  University of Illinois Urbana-Champaign\\
  \texttt{aadyaa3@illinois.edu} \\
  \And
  Ho Kei Cheng\\ %
  University of Illinois Urbana-Champaign\\
  \And 
    Alexander Schwing\\ %
  University of Illinois Urbana-Champaign\\
}

\begin{document}

\maketitle

\begin{figure}[h]
    \centering
    \includegraphics[width=\linewidth]{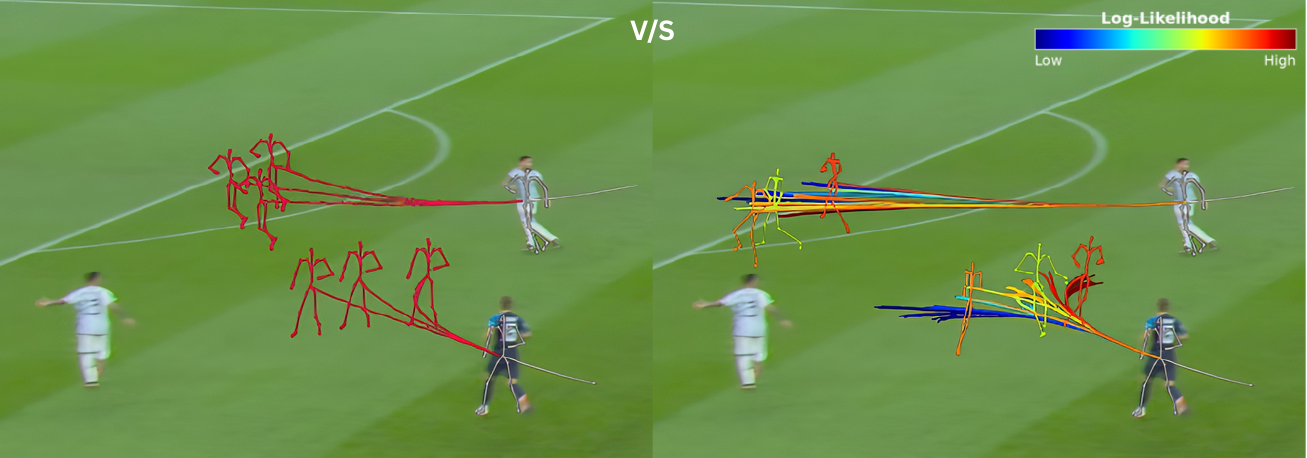}
    \caption{Motion forecasts projected onto broadcast video for two players. The gray skeleton and trails mark the recent motion history. (\emph{Left}) T2P deterministic predictions (red). (\emph{Right}) Our multi-modal predictions are colored by log-likelihood (red: high, blue: low).}
    \label{fig:teaser}
\end{figure}

\begin{abstract}

Predicting the future motion of multiple people is inherently stochastic: the same observed history admits many plausible continuations depending on intent, coordination, and social context. Yet the dominant paradigm in multi-person 3D motion prediction remains deterministic, with existing methods producing a fixed number of outputs per agent. 
In contrast, we present a conditional flow matching framework that jointly generates diverse future motions for multiple agents, modeling both global trajectories and root-relative poses. A B-spline reparameterization compresses the generative state space while enforcing temporal smoothness, and a dual-stream transformer conditions each agent's predictions on observed neighbor motion for interaction-aware forecasting. Our method achieves state-of-the-art performance on CMU-Mocap (UMPM) and 3DPW, outperforming deterministic baselines on joint position error, pose error, and final displacement error despite producing a full predictive distribution. We additionally establish the first motion prediction baseline on WorldPose, a large-scale professional soccer dataset, where social conditioning demonstrably concentrates predictive uncertainty along interaction-constrained directions. Our code is available at: \href{https://github.com/aadya-agrawal/FlowMoP}{https://github.com/aadya-agrawal/FlowMoP}.
\end{abstract}
\section{Introduction}

Anticipating how people move in shared environments underpins applications from autonomous navigation~\citep{paden2016survey} to sports analytics~\citep{li2021baseball}. The challenge is not just modeling individual body dynamics: people coordinate, react, and anticipate each other. Capturing these dependencies is essential for useful prediction. Recent work has made substantial progress through increasingly expressive interaction models~\citep{mrt2021, xu2023jrt, peng2023trajectory, zheng2025empmp}, yet a fundamental assumption persists across most prior works: future motion is treated as deterministic, and a fixed number of candidate trajectories are produced per agent.

Is a deterministic approach a good fit for the problem, given that human motion is inherently multi-modal? A person observed walking toward a junction could turn left, turn right, or stop, and the right answer depends on unobservable intent~\citep{GraberCVPR2020,GraberCVPR2020W}. %
Single-person stochastic approaches have addressed this through diffusion models~\citep{curreli2025nonisotropic} and masked autoencoders~\citep{humanmac}, but do not model inter-agent dependencies. While the multi-modal nature of human motion is increasingly acknowledged in evaluation~\citep{jeong2024multi}, extension to the multi-person setting remains underexplored, and existing models remain predominantly deterministic. Prior stochastic methods for multi-person 3D prediction rely on autoregressive rollout across short chunks~\citep{dummf} or focus on long-term interaction plausibility under custom realism metrics~\citep{socialdiffusion}. 
This leaves an open challenge for a unified generative framework. %

We address this challenge with a flow matching framework for multi-person 3D motion prediction. We decompose joint positions into a global root trajectory and root-relative local pose, modeling both within a shared generative model. Rather than generating every future frame independently, predictions are made in a compact B-spline control-point space, reducing integration cost while enforcing temporal smoothness. A dual-stream transformer processes the observed history and noisy future tokens through separate attention pathways, keeping past observations as stable conditioning throughout generation. For interaction-rich settings, neighbor trajectories are incorporated as additional context, conditioning each agent's predictive distribution on the observed motion of nearby agents.

We evaluate on standard multi-person benchmarks~\citep{vanderaa2011umpm, vonmarcard2018recovering} and establish, to our knowledge, the first motion prediction baseline on WorldPose~\citep{jiang2024worldpose}, a large-scale professional soccer dataset %
with 3D poses estimated from broadcast video rather than controlled capture.
Figure~\ref{fig:teaser} illustrates the benefit of generative modeling: our model exposes a distribution over plausible futures, whereas a deterministic baseline produces trajectory proposals whose corresponding pose predictions degenerate to near-static configurations.

\noindent Our key contributions are:
\begin{itemize}[leftmargin=*,nolistsep]
    \item A conditional flow matching framework for stochastic multi-person 3D motion prediction, jointly modeling global trajectory and local pose.
    \item A B-spline reparameterization of the prediction space that reduces integration cost and enforces temporal smoothness without sacrificing accuracy.
    \item A dual-stream transformer achieving state-of-the-art results on CMU-Mocap (UMPM) and 3DPW.
    \item An interaction study on WorldPose establishing the first generative baseline for full-scene soccer motion prediction under realistic broadcast conditions.
\end{itemize}
\section{Related Work}

\noindent\textbf{Multi-person 3D motion prediction.}
Early work on multi-person motion prediction focused on graph-based representations of skeleton interactions. MRT~\citep{mrt2021} introduced multi-range transformers with separate local and global encoders, capturing both individual dynamics and inter-person dependencies within a shared architecture. SoMoFormer~\citep{somoformer} extended this with joint-type and identity tokens, enabling finer-grained inference over body structure across agents. Subsequent work refined the interaction model: JRT~\citep{xu2023jrt} incorporated physical constraints through joint-relation graphs, TBIFormer~\citep{peng2023trajectory} partitioned the scene into body-part interaction tokens to capture contact and proximity cues, and T2P~\citep{jeong2024multi} introduced a coarse-to-fine formulation in which trajectory predictions condition pose generation. EMPMP~\citep{zheng2025empmp} achieves competitive performance with a lightweight MLP-based dual-branch architecture. Transformer forecasting has also been studied for a single person: 2CH-TR~\citep{valls2022robust} decouples spatial and temporal attention to forecast short- and long-term motion from a brief observation. All of these methods are deterministic.
Two prior works introduce stochasticity: DuMMF~\citep{dummf} uses a DDPM framework that relies on a small set of fixed learned motion intents to produce diverse predictions, inherently capping the variety of forecasts it can generate.
Social Diffusion~\citep{socialdiffusion} applies diffusion to multi-person 3D motion, focusing on long-term interaction plausibility under custom realism metrics. 
Our work is the first to apply flow matching to multi-person 3D motion prediction and to generate the full prediction horizon non-autoregressively.

\noindent\textbf{Stochastic single-person motion prediction.}
Stochastic single-person motion prediction has been explored through several generative frameworks. VAE-based methods~\citep{mld} learn compact latent representations of motion but are prone to posterior collapse. Diffusion-based approaches have shown strong results: SkeletonDiffusion~\citep{curreli2025nonisotropic} introduces non-isotropic noise schedules tailored to skeletal structure, and HumanMAC~\citep{humanmac} frames prediction as masked motion completion. Flow matching has been applied to text-conditioned motion generation~\citep{motionhiflow} and single-person motion prediction~\citep{cacheflow}, but not to multi-person 3D pose prediction. %
A separate line of work on human-human interaction generation~\citep{interx, intermask} synthesizes motion from text or action labels rather than predicting from observed history, and is complementary rather than directly comparable.

\noindent\textbf{Motion datasets and interaction richness.}
Progress in multi-person prediction has been shaped by the scale and diversity of available datasets. Laboratory capture datasets such as CMU-Mocap~\citep{cmu2003mocap}, UMPM~\citep{vanderaa2011umpm}, MuPoTS-3D~\citep{mupots}, and ExPI~\citep{expi} provide accurate 3D annotations but are limited to controlled environments with at most three subjects and scripted interactions. In-the-wild datasets such as 3DPW~\citep{vonmarcard2018recovering} and Harmony4D~\citep{harmony4d} extend to natural settings but remain limited to two-person scenarios. JRDBPose3D~\citep{jrdbpose3d} captures larger crowds but focuses on pedestrian navigation rather than purposeful coordinated behavior. WorldPose~\citep{jiang2024worldpose} addresses this gap with full 22-player professional soccer scenes, where interactions reflect strategic collective behavior. The scale and purposeful interaction structure of WorldPose motivate our interaction-aware conditioning and make it a compelling testbed for generative multi-person prediction.

\noindent\textbf{Trajectory reparameterization.}
Compact representations of motion trajectories have been explored in the past. DCT-based compression is used in deterministic pose prediction by SiMLPe~\citep{simlpe}, T2P~\citep{jeong2024multi}, and EMPMP~\citep{zheng2025empmp} to reduce sequence dimensionality in the frequency domain. Bézier curves have been applied to endpoint-conditioned trajectory prediction~\citep{pecnet} and motion planning. B-splines have recently been used for point tracking in video~\citep{traceanything}, where their $\mathcal{C}^2$ continuity and compact control-point representation offer advantages similar to those we exploit here. Our contribution is to apply clamped cubic B-splines to structured skeleton prediction within a flow matching model: we operate entirely in control-point space and project velocity residuals to full resolution via a fixed basis matrix for temporally-weighted supervision. To our knowledge, this is the first application of B-spline reparameterization to generative 3D motion prediction.
\section{Method}

\begin{figure}[t]
    \centerline{{\includegraphics[width=\linewidth]{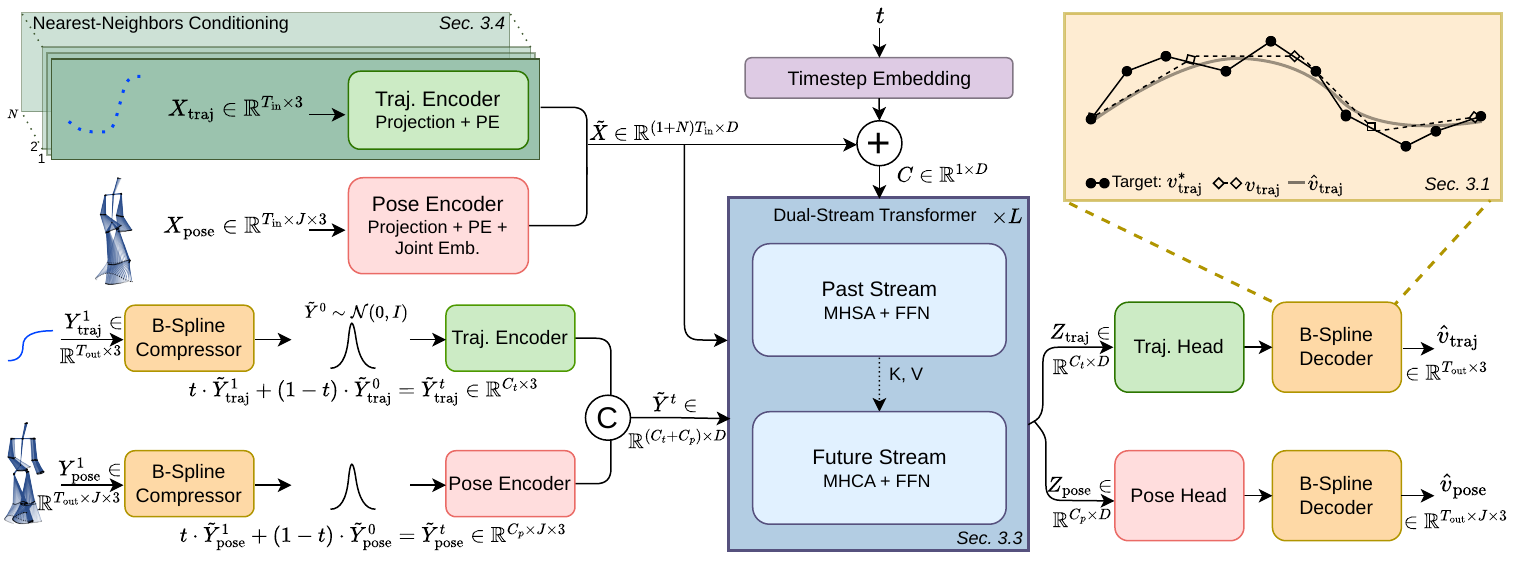}}}
    \vspace{-3pt}
    \caption{Overview of the proposed method. 
    \emph{(a)} Past trajectory and pose observations are encoded into a shared context $\tilde{\mathbf{X}}$. Noisy future control points are encoded and concatenated into $\tilde{\mathbf{Y}}^t$, conditioned on flow timestep $t$ via $\mathbf{C}$. 
    \emph{(b)} $L$ dual-stream transformer blocks process past and future tokens separately, with future tokens attending to $[\tilde{\mathbf{X}};\tilde{\mathbf{Y}}]$. 
    \emph{(c)} Velocity heads decode trajectory and pose control points, lifted to $T_\text{out}$ frames by B-spline evaluation.}
    \label{fig:arch}
    \vspace{-3pt}
\end{figure}

We predict the stochastic future motion of a focal agent over $T_\text{out}$ frames conditioned on $T_\text{in}$ observed frames. For this, we decompose absolute joint positions into 
a global trajectory $\mathbf{X}_\text{traj}$ 
tracking the root hip joint and a local pose $\mathbf{X}_\text{pose}$ 
of root-relative joint offsets for $J$ joints. Coordinates are expressed in an agent-centric frame 
translated to the last observed hip position without rotating the coordinate axes, rendering predictions invariant to absolute scene location. Given these observations, we learn to generate
the corresponding future trajectory $\hat{\mathbf{Y}}_\text{traj}$ and pose $\hat{\mathbf{Y}}_\text{pose}$. An overview of the full method is shown in Figure~\ref{fig:arch}, with the following subsections describing each component in turn. Table~\ref{tab:notation} summarizes the frame-space and control-point-space notation used.

\begin{table}[t]
    \centering
    \caption{Notation. Subscripts $q\in\{\mathrm{traj},\mathrm{pose}\}$ identify the branches, with $d_\mathrm{traj}=3$ and $d_\mathrm{pose}=3J$.}
    \label{tab:notation}
    \small
    \begin{tabularx}{\textwidth}{l l X}
    \toprule
    \textbf{Symbol} & \textbf{Shape} & \textbf{Meaning} \\
    \midrule
    $\mathbf{X}_q$ & $T_\mathrm{in}\times d_q$ & Observed sequence in frame space \\
    $\mathbf{Y}_q,\hat{\mathbf{Y}}_q$ & $T_\mathrm{out}\times d_q$ & Ground-truth and predicted future in frame space \\
    $\tilde{\mathbf{Y}}^0_q,\tilde{\mathbf{Y}}^1_q,\tilde{\mathbf{Y}}^t_q$ & $C_q\times d_q$ & Source noise, target, and interpolant in control-point space \\
    $\mathbf{B}_q,\mathbf{B}_q^\dagger$ & $T_\mathrm{out}\times C_q$, $C_q\times T_\mathrm{out}$ & Spline evaluation and projection matrices \\
    $v^*_q,\hat v_q$ & $T_\mathrm{out}\times d_q$ & Target and predicted velocities after projection to frame space \\
    \bottomrule
    \end{tabularx}
\end{table}

\subsection{B-Spline Control-Point Reparameterization}
\label{sec:spline}

We reparameterize predictions using a B-spline basis, compressing the future motion to $C < T_\text{out}$ control points while enforcing $\mathcal{C}^2$ temporal smoothness.

\noindent\textbf{Basis construction.}
For this, we construct a fixed basis matrix $\mathbf{B} \in \mathbb{R}^{T_\text{out} \times C}$ from a clamped cubic 
B-spline with uniformly spaced interior knots; full construction details are provided in Appendix~\ref{app:bspline}. 
Clamping ensures the spline passes exactly through its first and last control points \big($\mathbf{B}[0, :] = \mathbf{e}_1$, $\mathbf{B}[T_\text{out}-1, :] = \mathbf{e}_C$\big), so the final control point directly corresponds to the final predicted frame. Ground-truth control points are precomputed via the Moore-Penrose pseudoinverse as $\tilde{\mathbf{Y}}^1 = \mathbf{B}^\dagger\mathbf{Y}^1$, and full-resolution sequences are recovered as ${\mathbf{Y}} = \mathbf{B}\tilde{\mathbf{Y}}$. $\mathbf{B}$ is fixed throughout training, with gradients flowing through $\mathbf{B}^\top$ during backpropagation.

\noindent\textbf{Separate trajectory and pose compression.}
Global trajectory and local pose exhibit different temporal frequency content: root motion is globally smooth while joint articulations contain rapid high-frequency components. We therefore use independent basis matrices with $C_t$ control points for trajectory and $C_p > C_t$ control points for pose. %

\subsection{Conditional Flow Matching for Motion Prediction}
\label{sec:CFM}
Flow matching~\citep{lipman2022flow} learns a vector field $v_\theta$ transporting samples drawn from a source distribution $p_0$ to samples drawn from a target distribution $p_1$ via an ODE. 
We adopt the conditional variant with a linear interpolation path that operates entirely in control-point space. Ground-truth control points $\tilde{\mathbf{Y}}^1$ serve as the target, and noise is sampled directly in the same space: $\tilde{\mathbf{Y}}^0_\text{traj}, \tilde{\mathbf{Y}}^0_\text{pose} \sim \mathcal{N}(0, \mathbf{I})$. 
Sampling in control-point space, rather than frame space, preserves the isotropic Gaussian source distribution in the %
representation on which the ODE is defined.
Projecting frame-space noise through $\mathbf{B}^\dagger$ would otherwise introduce correlations between control points. Hence, our interpolant is:
\begin{equation}
    \tilde{\mathbf{Y}}^t = (1 - t)\tilde{\mathbf{Y}}^0 + t\tilde{\mathbf{Y}}^1.
\end{equation}

\noindent\textbf{Training.}
The model $v_\theta(\tilde{\mathbf{Y}}^t, t \mid \mathbf{X})$ is trained to predict the conditional velocity $\tilde{\mathbf{Y}}^1 - \tilde{\mathbf{Y}}^0$. During inference, full-resolution sequences are recovered via $\mathbf{B}$ only after integration; during training, velocities are projected to frame space for supervision. We sample $t$ from a logit-normal distribution~\citep{esser24scaling}.

\noindent\textbf{Inference.}
At inference, $K$ independent noise draws are each integrated with an Euler solver over $S$ steps. In some applications or for a fair comparison to prior work, $K' \ll K$ diverse representative predictions are required. For this, we apply k-means clustering in the space of last-frame joint positions, using trajectory coordinates up-weighted to balance their contribution with that of the higher-dimensional pose features. The $K'$ cluster medoids are returned as the final output, encouraging diversity without requiring an explicit diversity loss during training.

\noindent\textbf{Loss.}
Velocities are predicted in control-point space, and projected via $\mathbf{B}$ for temporally-weighted supervision. For each branch $q\in\{\mathrm{traj},\mathrm{pose}\}$, the frame-space target and prediction are $v^*_q=\mathbf{B}_q(\tilde{\mathbf{Y}}^1_q-\tilde{\mathbf{Y}}^0_q)$ and $\hat v_q=\mathbf{B}_q\hat v^\text{ctrl}_q$, respectively, where $\hat v^\text{ctrl}_q$ is the model output:
\begin{equation}
    \mathcal{L}_{\text{traj}} = \mathbb{E}_{t, \mathbf{Y}^0} \left[
        \bigl\| \mathbf{w}_t \odot (\hat{v}_{\text{traj}} - v^*_{\text{traj}})
        \bigr\|^2_2 \right], \quad
    \mathcal{L}_{\text{pose}} = \mathbb{E}_{t, \mathbf{Y}^0} \left[
        \bigl\| \mathbf{w}_p \odot (\hat{v}_{\text{pose}} - v^*_{\text{pose}})
        \bigr\|^2_2 \right]
    \label{eq:loss}
\end{equation}
where $\|\cdot\|_2^2$ denotes the mean squared error over frames, joints, and coordinates, and $\mathbf{w}_t$ and $\mathbf{w}_p$ increasingly weight later frames. Additionally, we supervise the approximate clean sample $\hat{\mathbf{Y}}^1=\mathbf{B}(\tilde{\mathbf{Y}}^t+\hat v^\text{ctrl}(1-t))$ with the final-frame joint-position loss:
\begin{equation}
    \mathcal{L}_\text{jpe}=\mathbb{E}_{t,\mathbf{Y}^0}\!\left[
    t^2\frac{1}{J}\sum_{j=1}^{J}
    \left\|\hat{\mathbf{Y}}^1_{T_\text{out},j}-
    \mathbf{Y}^1_{T_\text{out},j}\right\|_2\right].
\end{equation}
This directly matches the absolute JPE evaluation target, while $t^2$ emphasizes noise levels where the one-step clean estimate is most accurate. Exact weighting schedules and hyperparameters are provided in Appendix~\ref{app:implementation}. The overall loss is:
\begin{equation}
    \mathcal{L} = \lambda_\text{traj}\,\mathcal{L}_\text{traj} 
                + \lambda_\text{pose}\,\mathcal{L}_\text{pose}
                + \lambda_\text{jpe}\,\mathcal{L}_\text{jpe}.
\end{equation}
    
\subsection{Dual-Stream Transformer Architecture}
\label{sec:transformer_arch}

\noindent\textbf{Input encoding.}
Past trajectory tokens are formed 
by linearly projecting each frame of $\mathbf{X}_\text{traj}$ into $\mathbb{R}^D$ and adding a sinusoidal 
positional encoding.
Past pose tokens additionally use a structured per-joint encoding: each joint's 3D coordinates are projected and summed with a learned joint-type embedding $\mathbf{e}_j$ before being aggregated across joints into 
a single token per timestep. The two streams are then summed to yield the past context $\tilde{\mathbf{X}} \in \mathbb{R}^{T_\text{in} \times D}$.

Noisy future control points $(\tilde{\mathbf{Y}}^t_\text{traj}, \tilde{\mathbf{Y}}^t_\text{pose})$ are encoded symmetrically, with positional encoding (PE) evaluated at anchor frame indices $a_c = \lfloor (c/C) \cdot T_\text{out} \rfloor$ to preserve temporal alignment between the spline parameterization and the positional embedding. A learned modality-type embedding distinguishes trajectory from pose tokens before concatenation into $\tilde{\mathbf{Y}} \in \mathbb{R}^{(C_t + C_p) \times D}$.

\noindent\textbf{Conditioning.}
A sinusoidal timestep embedding is transformed by an MLP and summed with the mean-pooled past context to form $\mathbf{C} \in \mathbb{R}^{1 \times D}$. A shared linear projection of $\mathbf{C}$ produces eight adaLN modulation vectors per block (scales and shifts for each of: past attention, past MLP, future attention, future MLP).

\noindent\textbf{Dual-stream blocks.}
Over $L$ blocks, the past stream performs self-attention over the past context $\tilde{\mathbf{X}}$ only. Its encoded input is fixed across depth: each block re-reads the same $\tilde{\mathbf{X}}$. Within the block past tokens are modulated by self-attention, an MLP, and their own adaLN scale and shift. The future stream attends to the complete concatenated context $[\tilde{\mathbf{X}};\tilde{\mathbf{Y}}]$, allowing every future token to re-anchor to the unchanged observations at every block. Both streams have independent MLP sublayers and adaLN pre-norm, and per-head RMS normalization is applied for training stability.

\noindent\textbf{Output heads.}
After $L$ blocks, future tokens are partitioned into $\mathbf{Z}_\text{traj} \in \mathbb{R}^{C_t \times D}$ and $\mathbf{Z}_\text{pose} \in \mathbb{R}^{C_p \times D}$, each decoded by a linear head to velocity control points $\hat{v}_\text{traj} \in \mathbb{R}^{C_t \times 3}$ and 
$\hat{v}_\text{pose} \in \mathbb{R}^{C_p \times J \cdot 3}$.

\subsection{Nearest-Neighbor Conditioning}
\label{sec:interaction}

For datasets with rich multi-agent interactions, such as team sports~\citep{jiang2024worldpose}, where player motion is tightly coupled conditioning on neighboring agents. To this end, we optionally condition each focal agent's prediction on the $N$ nearest neighbors according to Euclidean distance at the last observed frame. Each neighbor's past root trajectory is expressed in the focal agent's coordinate frame by subtracting the focal agent's last observed position, encoding relative spatial relationships rather than absolute positions.
We adopt a per-agent formulation conditioned on neighbor observations, keeping the problem tractable across variable-size scenes and allowing the model to be trained and evaluated independently per agent. Social awareness is preserved: each agent's predictive distribution is explicitly conditioned 
on observed neighbor motion, and our results confirm this is sufficient for accurate per-agent prediction. Extending the model to joint prediction could enforce mutual consistency across all agents' futures. This remains future work.

Neighboring trajectories are encoded with the same linear projection as the ego trajectory, with a learned agent-type embedding added to distinguish them. The resulting $N \cdot T_\text{in}$ neighbor tokens are prepended to the ego's past context, extending $\tilde{\mathbf{X}}$ to $\mathbb{R}^{(1+N)T_\text{in} \times D}$. For scenes with fewer than $N$ agents, invalid neighbor tokens are set to zero via a validity mask broadcast across the temporal dimension, enabling batched training across scenes with variable occupancy.
\section{Experiments}
\label{sec:exp}
\subsection{Setup}
\label{sec:exp:setup}
\noindent\textbf{Datasets.}
We evaluate on three 3D multi-person benchmarks, following the protocol of~\citet{jeong2024multi} for CMU-Mocap (UMPM) and 3DPW to enable direct comparison with existing methods.
CMU-Mocap (UMPM)~\citep{cmu2003mocap, vanderaa2011umpm} contains motion capture sequences of 3 agents with 15 joints, using 1.0s of observation for 2.0s of prediction. 3DPW~\citep{vonmarcard2018recovering} contains in-the-wild sequences of 2 agents with 13 joints, using 0.8s of observation for 1.6s of prediction. WorldPose~\citep{jiang2024worldpose} is a large-scale professional soccer dataset with full 22-player scenes and 24 SMPL joints per agent, with 3D poses estimated from broadcast video. We predict 3.0s of motion from 1.0s of observation. WorldPose results are discussed separately in \S\ref{sec:worldpose_results}.

\noindent\textbf{Metrics.}
We report Joint Position Error (JPE), Average Pose Error (APE), and Final Displacement Error (FDE), all in millimeters. JPE measures absolute joint error, APE measures root-relative pose error, and FDE measures root error at the final predicted frame. Following~\citet{jeong2024multi}, we generate $K'=6$ diverse predictions per agent and report best-of-$K'$ metrics. Formal definitions are provided in appendix~\S\ref{app:metrics}.

\noindent\textbf{Baselines.}
We compare against six multi-person motion prediction methods. MRT~\citep{mrt2021} models inter-person dependencies via a multi-person transformer with cross-attention between agents. JRT~\citep{xu2023jrt} extends this with a joint-level relational transformer that explicitly reasons over individual joints across agents. TBIFormer~\citep{peng2023trajectory} partitions the scene into body-part interaction tokens, capturing fine-grained inter-person contact and proximity cues. T2P~\citep{jeong2024multi} adopts a two-stage approach, sequentially predicting root trajectories and then conditioning pose prediction on them. EMPMP~\citep{zheng2025empmp} is the current state of the art, using a graph-based interaction model over agent skeletons. We additionally compare against DuMMF~\citep{dummf}, a DDPM-based stochastic method that generates diverse multi-person predictions autoregressively via a dual-level transformer denoiser trained with individual and social interaction objectives.

\noindent\textbf{Implementation details.}
The transformer uses $L=8$ blocks with embedding dimension $D=256$. We use $C_t=8$ and $C_p=24$ B-spline control points for the trajectory and pose, respectively. Training uses AdamW with learning rate $1\times10^{-4}$ and a cosine schedule for $20k-50k$ steps, with logit-normal timestep sampling ($\mu_t=-0.5$, $s_t=1.5$). Past frames are randomly dropped from the start of the input sequence with probability 0.3 during training. Inference uses $S=10$ Euler steps with $K=100$ proposals downselected to $K'=6$ via weighted k-means.

\subsection{Results: 3D Multi-Person Benchmarks}

\begin{table}[t]
    \centering
    \caption{Comparison of best-of-6 JPE, APE, and FDE metrics (in \emph{mm}) at different time horizons on CMU-Mocap (UMPM) and 3DPW datasets. Lower values are better $(\downarrow)$, with the best results shown in \textbf{bold}. An asterisk $(^*)$ denotes models we recomputed for this setup.}
    \label{tab:motion_eval_consolidated}
    \small
    \setlength{\tabcolsep}{3.5pt} 
    \resizebox{\textwidth}{!}{%
    \begin{tabular}{llc cccc cccc cccc}
    \toprule
    & & & \multicolumn{6}{c}{\textbf{CMU-Mocap (UMPM)}} & \multicolumn{6}{c}{\textbf{3DPW}} \\
    \cmidrule(lr){4-9} \cmidrule(lr){10-15}
    & & & \multicolumn{2}{c}{\textbf{JPE ($\downarrow$)}} & \multicolumn{2}{c}{\textbf{APE ($\downarrow$)}} & \multicolumn{2}{c}{\textbf{FDE ($\downarrow$)}} & \multicolumn{2}{c}{\textbf{JPE ($\downarrow$)}} & \multicolumn{2}{c}{\textbf{APE ($\downarrow$)}} & \multicolumn{2}{c}{\textbf{FDE ($\downarrow$)}} \\
    \cmidrule(lr){4-5} \cmidrule(lr){6-7} \cmidrule(lr){8-9} \cmidrule(lr){10-11} \cmidrule(lr){12-13} \cmidrule(lr){14-15}
    \textbf{Model} & \textbf{Venue} & \textbf{Gen.?} & \textbf{1.0s} & \textbf{2.0s} & \textbf{1.0s} & \textbf{2.0s} & \textbf{1.0s} & \textbf{2.0s} & \textbf{0.8s} & \textbf{1.6s} & \textbf{0.8s} & \textbf{1.6s} & \textbf{0.8s} & \textbf{1.6s} \\
    \midrule
    MRT       & NeurIPS\textquotesingle21 & \color{red}{\ding{55}} & 164.7 & 280.1 & 127.0 & 164.4 & 99.6  & 204.7 & 159.1 & 251.2 & 117.9 & 153.2 & 102.7 & 185.3 \\
    DuMMF*  & ICLR\textquotesingle23 & \color{green}{\ding{51}}  & 162.9 & 275.0 & 125.3 & 161.1 & 97.6 & 199.2 & 158.5 & 248.7 & 115.5 & 151.9 & 102.8 & 183.1 \\
    TBIFormer & CVPR\textquotesingle23    & \color{red}{\ding{55}} & 170.0 & 290.9 & 125.1 & 160.8 & 112.1 & 228.5 & 153.9 & 265.8 & 115.4 & 152.7 & 106.7 & 215.9 \\
    JRT       & ICCV\textquotesingle23    & \color{red}{\ding{55}} & 168.5 & 316.9 & 121.2 & 181.6 & 117.7 & 250.8 & 181.9 & 287.3 & 133.4 & 178.0 & 133.7 & 235.4 \\
    T2P       & CVPR\textquotesingle24    & \color{red}{\ding{55}} & 152.4 & 262.7 & 114.4 & 151.7 & 88.7  & 188.9 & 142.6 & 236.2 & 114.6 & 150.0 & 74.1  & 158.2 \\
    EMPMP$^*$ & ICCV\textquotesingle25    & \color{red}{\ding{55}} & 146.2 & 250.4 & 100.8 & 146.5 & 85.5  & 171.6 & 140.3 & 235.4 & 111.9 & 150.6 & 71.8  & 153.0 \\
    \textbf{Ours} &      & \color{green}{\ding{51}} & \textbf{132.3} & \textbf{237.4} & \textbf{92.4} & \textbf{128.6} & \textbf{81.2} & \textbf{159.3} & \textbf{139.3} & \textbf{232.1} & \textbf{109.7} & \textbf{146.8} & \textbf{69.2} & \textbf{150.5} \\
    \bottomrule
    \end{tabular}%
    }
\end{table}

Table~\ref{tab:motion_eval_consolidated} reports results on CMU-Mocap (UMPM) and 3DPW. Our approach achieves state-of-the-art performance across all metrics. On CMU-Mocap, we improve by 12.2\% on APE and 7.2\% on FDE against the strongest deterministic baseline, EMPMP. Notably, against the generative model DuMMF, we achieve a substantial improvement of 13.7\% on JPE and 20.2\% on APE at 2.0s. This suggests that our continuous-time ODE, which generates the entire future horizon in a single pass, effectively avoids the cumulative error and temporal drift inherent in DuMMF's autoregressive rollout strategy.
The improvement is most pronounced on APE across both datasets, reflecting the benefit of the structured per-joint encoding and separate trajectory-pose decomposition for local pose accuracy. Our model also outperforms previous methods across all six 3DPW metrics, demonstrating its effectiveness beyond the controlled capture setting.

\begin{figure}[t]
    \centering
    \includegraphics[width=\linewidth]{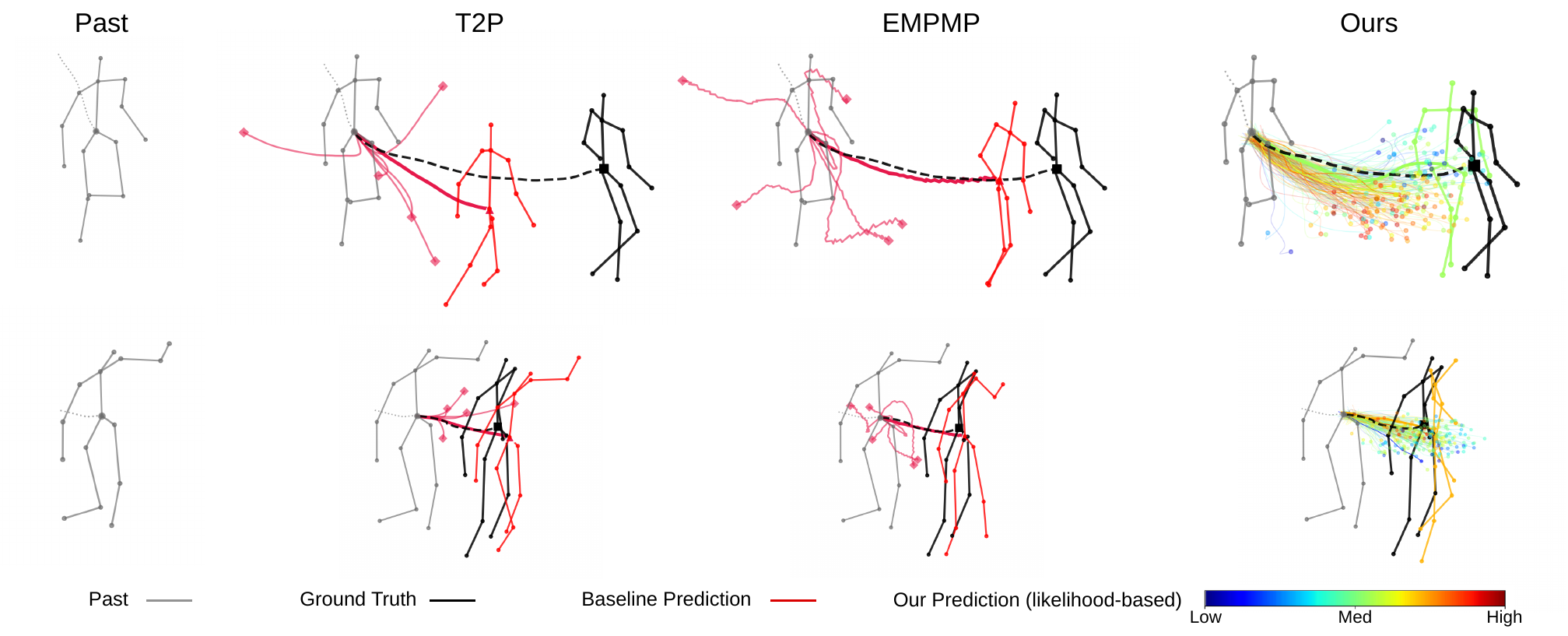}
    \vspace{-8mm}
    \caption{Qualitative comparison on CMU-Mocap (UMPM). Each column shows the root trajectory and final-frame skeleton for a single test agent: the observed past, baselines, and our model. Dashed black lines denote ground truth future trajectory; colored lines show the trajectories of predictions.}
    \label{fig:density_map}
\end{figure}
Figure~\ref{fig:density_map} illustrates a key benefit of our generative model. Visualizing all 100 proposals, colored by log-likelihood, reveals how probability mass concentrates around the model's most likely motion modes, showing a rich learned prior over plausible futures. Deterministic baselines produce a deterministic prediction. %
Differently, our model  exposes the full predictive distribution, enabling downstream selection or likelihood-aware planning.

\subsection{Interaction Study on WorldPose}
\label{sec:worldpose_results}

\begin{table}[t]
    \centering
    \small
    \setlength{\tabcolsep}{9pt}
    \caption{Comparison of best-of-6 JPE, APE, and FDE metrics (in \emph{m}) at different time horizons on WorldPose soccer data. Lower values are better $(\downarrow)$, with the best results shown in \textbf{bold}. An asterisk $(^*)$ denotes models we recomputed for this setup.}
    \label{tab:worldpose_maintable}
    \small
    \begin{tabular}{l ccc ccc ccc}
    \toprule
    & \multicolumn{3}{c}{\textbf{JPE ($\downarrow$)}} & \multicolumn{3}{c}{\textbf{APE ($\downarrow$)}} & \multicolumn{3}{c}{\textbf{FDE ($\downarrow$)}} \\
    \cmidrule(lr){2-4} \cmidrule(lr){5-7} \cmidrule(lr){8-10}
     \textbf{Model} & \textbf{1.0s} & \textbf{2.0s} & \textbf{3.0s} & \textbf{1.0s} & \textbf{2.0s} & \textbf{3.0s} & \textbf{1.0s} & \textbf{2.0s} & \textbf{3.0s} \\
    \midrule
    T2P$^*$  & 0.91 & 1.70 & 1.99 & 0.19 & 0.22 & 0.26 & 0.84 & 1.50 & 1.77\\
    EMPMP$^*$ &  0.92 & 1.79 & 1.96 & 0.15 & 0.16 & 0.19 & 0.86 & 1.64 & 1.93\\
    \textbf{Ours} w. $N=0$ &  0.87 & 1.41 & 1.87 & 0.08 & 0.09 & 0.10 & 0.80 & 1.34 & 1.79\\
    \textbf{Ours} w. $N=3$ &  0.83 & 1.37 & 1.84 & 0.07 & 0.09 & 0.10 & 0.78 & 1.32 & 1.77\\
    \textbf{Ours} w. $N=10$ & \textbf{0.80} & \textbf{1.36} & \textbf{1.82} & \textbf{0.06} & \textbf{0.09} & \textbf{0.10} & \textbf{0.77} & \textbf{1.31} & \textbf{1.75}\\
    \bottomrule
    \end{tabular}
\end{table}

WorldPose presents a substantially harder setting than CMU-Mocap (UMPM) and 3DPW: player motion in soccer is governed by strategic interactions, including positioning, pressing, and off-ball runs that unfold over longer horizons and involve many more agents simultaneously. This makes it an ideal testbed for interaction-aware conditioning at scale.

Table~\ref{tab:worldpose_maintable} reports results across 1.0--3.0s horizons. Even without neighbor conditioning $(N=0)$, our method improves upon T2P and EMPMP, with particularly large gains on APE, highlighting the benefits of the generative formulation and structured pose encoding, independent of interaction modeling. Incorporating neighbors yields consistent further improvements across all metrics and horizons, with gains increasing with the number of neighbors. The improvement is most pronounced on JPE and FDE at longer horizons, where strategic positioning becomes increasingly influential: knowing where nearby players are heading constrains the focal agent's plausible futures. The diminishing returns between $N=3$ and $N=10$ suggest that the most informative social context comes from a small number of nearby agents, with additional neighbors contributing marginally.

\begin{figure}[t]
    \centering
    \includegraphics[width=\linewidth]{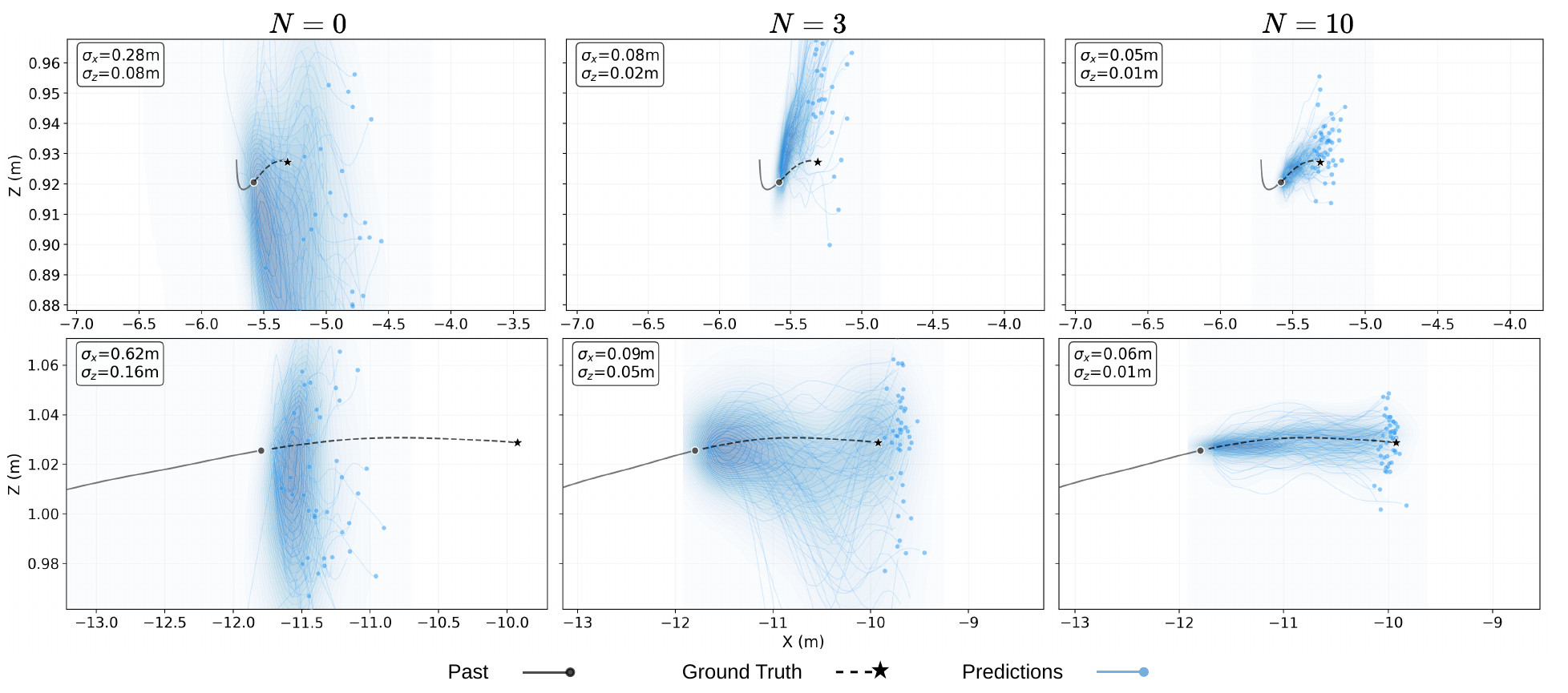}
    \caption{Predicted hip trajectory distributions under varying social contexts for two players. Each panel shows 50 predicted root trajectories with a KDE density overlay, the observed past trajectory (gray), the ground-truth future path (dashed black), and predicted endpoints (scatter).}
    \label{fig:worldpose_qual2}
\end{figure}

Figure~\ref{fig:worldpose_qual2} shows predicted root trajectories for representative agents under the three neighbor conditioning settings. Without social context $(N=0)$, the model produces a broad, high-variance distribution, reflecting reliance on learned motion priors. Conditioning on neighbors increasingly concentrates predictions along the ground-truth direction, converging into a tight forward-directed fan.

This compression is axis-asymmetric, with $\sigma_x / \sigma_z$ increasing from $\sim 3.5$ to $\sim 6$ with 10 neighbors considered, indicating that the model learns real interaction constraints along the axis of collective motion, not just reducing overall variance. Forward motion is most constrained by social context, as neighbors provide strong cues about the direction of play, while residual lateral spread reflects positional ambiguity that persists even with complete social information. The predicted density peak aligns with the ground truth under neighbor conditioning, demonstrating that social context serves as an effective implicit prior over plausible motion directions.

\section{Ablation Studies}
\label{sec:ablation}

All ablation experiments are conducted on CMU-Mocap (UMPM) and report best-of-6 JPE and APE at the 2.0s horizon. %

\begin{table}[t]
    \centering
    \caption{Ablation study of temporal representations. JPE and APE are reported in \emph{mm} at 2.0s horizon. Smoothness is measured via NPSS $\in [0,1]$. Forward Pass Latency is the GPU wall-clock time for a single forward call with batch size 64, mean$\pm$std. of 50 runs in \emph{ms}. Best results are shown in \textbf{bold}.}
    \label{tab:representation_comparison}
    \small
    \resizebox{\textwidth}{!}{%
    \begin{tabular}{ll cc cccc}
    \toprule
    \textbf{Representation} & \textbf{Config.} & \textbf{Traj. Basis} & \textbf{Pose Basis} & \textbf{JPE ($\downarrow$)} & \textbf{APE ($\downarrow$)} & \textbf{Smoothness ($\uparrow$)} & \textbf{Fwd. Pass Latency ($\downarrow$)}\\
    \midrule
    Frame-based & Raw & 50 & 50 & 245.8 & 133.3 & 0.62 & 7.86 $\pm$ 0.22\\
    \midrule
    Freq.-based & DCT & 50 & 50 &  249.7 & 133.7 & 0.61 & 12.30 $\pm$ 0.49 \\
    \midrule
    \multirow{5}{*}{Spline-based} & \multirow{2}{*}{Linear} & 4 & 12 & 238.8 & 129.2 & 0.67 & \textbf{6.41 $\pm$ 0.31}\\
    & & 8 & 24 &  239.4 & 129.2 & 0.67 & 6.59 $\pm$ 0.32\\
    \cmidrule(lr){2-8}
    & \multirow{3}{*}{Cubic} & 4 & 12 & 237.5 & 129.0 & \textbf{0.68} & 6.46 $\pm$ 0.16\\
    & & 8 & 24 &  \textbf{237.4} & \textbf{128.6}  & 0.67 & 6.61 $\pm$ 0.40  \\
    & & 12 & 36 & 241.7 & 129.8 & 0.66 & 7.17 $\pm$  0.18\\
    \bottomrule
    \end{tabular}%
    }
    \vspace{-10pt} 
\end{table}

\noindent\textbf{Temporal Representation.}
We compare three families of temporal basis for representing the future motion sequence within the ODE: raw frame-by-frame prediction, frequency-space DCT coefficients, and B-spline control points (Table~\ref{tab:representation_comparison}). All variants use the same transformer backbone; only the output representation and loss target change. 

Raw frame-based prediction and DCT perform comparably, with DCT offering no advantage despite its widespread use in deterministic motion forecasting~\citep{jeong2024multi, zheng2025empmp}, and incurring higher forward pass latency due to basis transformation overhead. Both spline variants outperform the frame-based baselines on all metrics, including smoothness as measured by NPSS~\citep{gopalakrishnan2019neural}, confirming that the structured inductive bias of $\mathcal{C}^2$ continuity benefits both accuracy and perceptual realism. Cubic splines consistently outperform piecewise-linear ones at matched control-point counts, justifying the added complexity of the Cox-de Boor recursion.

Within cubic splines, the choice of control-point count reveals an interesting trade-off: too few control points ($C_t{=}4, C_p{=}12$) slightly under-expresses fast joint dynamics, while too many ($C_t{=}12, C_p{=}36$) approaches the redundancy of the raw frame representation and loses the regularization benefit of compression. The intermediate setting ($C_t{=}8, C_p{=}24$) achieves the best JPE and APE, and reduces forward pass latency to 6.6ms relative to the 7.9ms raw frame baseline, a consequence of processing $C_t + C_p = 32$ future tokens through attention rather than $T_\text{out} = 50$.

\noindent\textbf{Architecture Components.} 
Table~\ref{tab:arch_ablation} traces the contribution of each architectural decision. Keeping past tokens fixed throughout the transformer enforces a clean information asymmetry: 
\begin{wraptable}{r}{0.5\textwidth}
    \centering
    \vspace{-12pt}
    \caption{Ablation study of architecture components. Each row removes or replaces a single component independently. $\Delta$ reported at 2.0s in \emph{mm}.}
    \label{tab:arch_ablation}
    \small
    \begin{tabular}{l cc}
    \toprule
    \textbf{Component change} & $\boldsymbol{\Delta}$\textbf{JPE} & $\boldsymbol{\Delta}$\textbf{APE} \\
    \midrule
    \multicolumn{3}{l}{\textit{Input encoding}} \\
    w/o per-joint encoding                         & +6.4 & +8.4 \\
    sep. $\rightarrow$ shared traj/pose tokens     & +18.2 & +11.9 \\
    \multicolumn{3}{l}{\textit{Attention}} \\
    dual $\rightarrow$ single-stream               & +12.8 & +5.3 \\
    fixed $\rightarrow$ iterative past context     & +19.9 & +7.8 \\
    \multicolumn{3}{l}{\textit{Conditioning}} \\
    adaLN $\rightarrow$ standard LayerNorm         & +5.1 & +6.5 \\
    w/o QK-norm                                    & +2.7 & +2.2 \\
    \midrule
    \textbf{Ours}                            & \textbf{237.4} & \textbf{128.6} \\
    \bottomrule
    \end{tabular}
    \vspace{-15pt} 
\end{wraptable}
past observations condition the future stream without being modified by it. This also acts as an implicit regularizer, reducing the model's capacity to overfit to the noisy intermediate states $\tilde{\mathbf{Y}}^t$ encountered during ODE integration.

Maintaining separate token streams for trajectory and pose is critical: collapsing the two into a single token per timestep discards structure that the transformer cannot recover, confirming that the decomposition must be preserved through the encoding stage, not just at the input and output levels.

The dual-stream attention builds on this foundation: future tokens attend to the concatenated $[\tilde{\mathbf{X}};\tilde{\mathbf{Y}}]$ context at every layer, continuously re-anchoring each prediction to the observed past as integration progresses. Per-joint encoding provides an additional gain, confirming that preserving joint identity in the input encoding is a useful inductive bias for pose accuracy. AdaLN conditioning and QK-norm contribute modestly; the latter primarily stabilizes training rather than improving accuracy, consistent with its role as a numerical stabilizer in flow matching~\citep{henry-etal-2020-query}.

\noindent\textbf{Training Components.} 
Table~\ref{tab:training_ablation} evaluates training decisions progressively, starting from a flow matching baseline with uniform timestep sampling, control-point-space loss, and no auxiliary losses or regularization (row 1). The largest single gain comes from $\mathcal{L}_\text{jpe}$ ($\sim$6.3mm JPE), which provides complementary supervision in absolute joint space beyond the decomposed trajectory and pose losses. Computing the velocity loss in frame space by projecting the control-point residual through $\mathbf{B}$ before applying temporal weights yields the next largest gain ($\sim$5.8mm JPE), suggesting that it is beneficial to distribute the gradient signal uniformly across all $T_\text{out}$ output frames than supervision only at the $C$ compressed coordinates. Context drop and data augmentation each provide further gains. Logit-normal timestep sampling contributes a smaller but consistent improvement, consistent with findings in image generation that concentrating training probability on intermediate noise levels is more efficient than uniform sampling~\citep{esser24scaling}.

\newcolumntype{Y}{>{\centering\arraybackslash}X}
\begin{table}[t]
    \centering
    \caption{Ablation study of training components. JPE and APE are reported in \emph{mm} at 2.0s horizon.}
    \label{tab:training_ablation}
    \small
    \begin{tabularx}{\textwidth}{l YYYYY cc}
    \toprule
    \textbf{ID} & \textbf{Logit-normal $t$} & \textbf{Frame-space loss} & \textbf{$\mathcal{L}_\text{jpe}$} & \textbf{Context drop} & \textbf{Data aug.} & \textbf{JPE ($\downarrow$)} & \textbf{APE ($\downarrow$)} \\
    \midrule
    (1) & & & & & & 256.5 & 142.2 \\
    (2) & \checkmark & & & & & 255.0 & 141.3 \\    
    (3) & \checkmark & \checkmark & & & & 249.2 & 138.8 \\
    (4) & \checkmark & \checkmark & \checkmark & & & 242.9 & 134.7 \\
    (5) & \checkmark & \checkmark & \checkmark & \checkmark & & 238.1 & 130.0 \\
    (6) & \checkmark & \checkmark & \checkmark & \checkmark & \checkmark & \textbf{237.4} & \textbf{128.6} \\
    \bottomrule
    \end{tabularx}
\end{table}

\section{Conclusion}

We presented a flow-matching framework for stochastic 3D human motion prediction that jointly models global trajectories and root-relative poses and achieved 
state-of-the-art performance on CMU-Mocap (UMPM) and 3DPW, outperforming baselines despite producing a full predictive distribution rather than a single output. On WorldPose, we showed 
evidence that the model learns genuine interaction constraints rather than simply reducing overall variance. 

A valuable direction for future work is efficient joint prediction over all agents simultaneously, which would allow the model to enforce mutual consistency across predicted trajectories. Within team-sport settings more broadly, incorporating ball trajectories could provide a shared reference that mediates interactions among all players that is not captured by pairwise neighbor conditioning alone.
For longer horizons, jointly increasing the control-point budget avoids recursive rollout, while adaptive counts or non-uniform knots may better preserve abrupt high-curvature motion; both remain to be validated beyond our 3.0s setting.
Learning explicit interaction patterns, such as pressing, off-ball runs, and defensive shape formation from the predictive distribution, also opens up opportunities for tactical analysis.

\noindent\textbf{Acknowledgements.} Work supported in part by NSF grants 2008387, 2045586, 2106825, 2551476, MRI 1725729, and NIFA award 2020-67021-32799.

\newpage
\bibliographystyle{plainnat}
\bibliography{neurips_2026}

\newpage
\appendix
\section*{Appendix}
The appendix is organized as follows. 
Appendix~\ref{app:implementation} provides additional implementation details covering dataset preparation, data augmentation, architecture hyperparameters, training configuration, loss scales, inference procedure, k-means downselection, evaluation metric definitions, and baseline retraining procedures. 
Appendix~\ref{app:bspline} gives the full B-spline basis construction, including the clamped knot vector, Cox-de Boor recursion, and basis matrix derivation. 
Appendix~\ref{app:flow} presents the flow matching training and inference algorithms, along with the likelihood scoring procedure used for visualizations. 
Appendix~\ref{app:ablation} reports additional ablation studies across datasets. 
Appendix~\ref{app:appendix_diversity} details our diversity evaluation with our generative baseline.
Appendix~\ref{app:qual} provides additional qualitative results on 3DPW and WorldPose. 
Appendix~\ref{app:limitation} discusses limitations and failure cases. 
Finally, Appendix~\ref{app:impacts} discusses the broader impact of our work, followed by license information in Appendix~\ref{app:licenses}.

\section{Additional Implementation Details}
\label{app:implementation}
\paragraph{Dataset.}
We train and evaluate on the UMPM split of the CMU-Mocap dataset~\citep{vanderaa2011umpm}, which provides multi-person indoor motion sequences captured at 25 FPS, and on 3DPW~\citep{vonmarcard2018recovering}, which provides in-the-wild two-person sequences at 12.5 FPS. Both follow the data preparation and processing pipeline of T2P~\citep{jeong2024multi}. For CMU-Mocap (UMPM), each person is represented by a 15-joint skeleton with the kinematic tree rooted at the hip joint. We observe $T_\text{in}=25$ frames (1.0s) and predict $T_\text{out}=50$ frames (2.0s). For 3DPW, a skeleton has 13 joints, and we predict $T_\text{out}=20$ frames from $T_\text{in}=10$ historical frames.

During training, we apply the following on-the-fly augmentations per sample: random scene rotation about the vertical axis, vertical reflection, isotropic scaling ($\pm8\%$), random permutation of agent order, and additive coordinate noise ($\sigma=5\,\text{mm}$). Skeleton mirroring is applied with probability 0.5 by swapping paired left and right joint indices and negating the lateral ($x$) axis; Additionally, past frames are randomly dropped from the beginning of the input sequence with probability 0.3, encouraging robustness to partial observations.

\paragraph{Architecture.}
The transformer uses $L=8$ dual-stream blocks with an embedding dimension $D=256$, $H=4$ attention heads, and an MLP expansion ratio of 2. A dropout of 0.1 is applied within each block. Per-head RMS normalization (QK-norm) is applied to queries and keys before attention. The timestep embedding has dimension $D=256$ and is transformed by a two-layer MLP with SiLU activations before being combined with the mean-pooled past context to form the conditioning vector $\mathbf{C} \in \mathbb{R}^{1 \times D}$. We use $C_t=8$ B-spline control points for trajectory and $C_p=24$ for pose, giving a total future token sequence of length $C_t + C_p = 32$, compared to $T_\text{out}=50$ in the uncompressed setting.

\paragraph{WorldPose dataset processing.}
We additionally evaluate on the WorldPose football dataset~\citep{jiang2024worldpose}, which provides SMPL-estimated 3D joint positions for up to 22 players per match at 25 Hz. Raw axis-angle parameters are converted to world-space 3D coordinates via SMPL forward kinematics, yielding 24-joint pose sequences. Sequences are extracted with a sliding window of stride 12 frames (0.5\,s); windows with more than 50\% missing detections are discarded, and shorter gaps ($\leq 5$ frames) are forward-filled. We predict $T_\text{out}=75$ future frames (3.0\,s) from $T_\text{in}=25$ observed frames (1.0\,s), with splits held out at the match level (80\% train / 20\% test, fixed seed). Augmentations follow the CMU-Mocap (UMPM) setting.

The longer horizon and faster, more erratic motion of football players require greater B-spline expressiveness than indoor MoCap. We use $C_t=30$ trajectory and $C_p=45$ pose control points, compared to $C_t=8$ and $C_p=24$ for CMU-Mocap. The higher counts are necessary because the output horizon is 50\% longer, and football motion involves sharp directional changes and explosive accelerations that a coarser basis cannot represent without significant reconstruction error. To verify this, we measured the mean spline reconstruction error by projecting ground-truth sequences to control points via $\mathbf{B}^\dagger$ and evaluating $\|\mathbf{B}\mathbf{B}^\dagger\mathbf{Y} - \mathbf{Y}\|$. The $(C_t=8, C_p=24)$ configuration incurs approximately $3\times$ higher JPE reconstruction error on WorldPose sequences than the $(C_t=30, C_p=45)$ configuration, justifying the increase.

\paragraph{Training.}
We use AdamW with $\beta=(0.9,\,0.999)$ and weight decay $10^{-2}$. Gradients are clipped to unit norm. The learning rate is $1\times10^{-4}$ and follows a cosine schedule with 6 warm restart cycles and a 200-step linear warmup at the start of training. An EMA of the model weights with decay $0.9995$ is maintained throughout training and used for all evaluation and checkpointing. 
We train with batch size 64 on a single NVIDIA L40S GPU (48\,GB VRAM). Each main model training run takes $\sim$2 hours (300 epochs). Ablation experiments collectively required approximately 100 GPU-hours, including the temporal representation sweep (7 variants), architecture component ablations (6 variants), training component ablations (6 variants), and the appendix ablations (7 variants). Baseline retraining for EMPMP, DuMMF, and T2P on WorldPose required approximately 40 additional GPU-hours.

\paragraph{Loss.}
We set the loss weights to $\lambda_\text{pose}=5.0$ and $\lambda_\text{jpe}=0.5$.
The total loss is therefore
\begin{equation}
    \mathcal{L} = \,\mathcal{L}_\text{traj} + 5.0\,\mathcal{L}_\text{pose}  + 0.5\,\mathcal{L}_\text{jpe}, 
\end{equation}
where $\mathcal{L}_\text{traj}$ and $\mathcal{L}_\text{pose}$ are the flow-matching velocity losses projected to full resolution via the spline basis $\mathbf{B}$, as defined in Eq.~\ref{eq:loss}. Both are temporally weighted with an exponential ramp $w_i \propto \exp(\alpha \cdot i / T_\text{out})$, normalized to unit mean, with $\alpha=5$ for trajectory (placing strong emphasis on final-frame displacement) and $\alpha=2$ for pose. 
$\mathcal{L}_\text{jpe}$ supervises absolute joint position error at the final frame, computed on the approximate clean sample $\hat{\mathbf{Y}}^1 \approx \tilde{\mathbf{Y}}^t + \hat{v}^\text{ctrl} \cdot (1-t)$, projected to full resolution and weighted by $t^2$ to concentrate supervision near $t=1$.

\paragraph{Inference.}
Trajectory and pose noise are sampled independently from $\mathcal{N}(0, \mathbf{I})$ in control-point space. 
For each agent, $K=100$ proposals are generated using $S=10$ Euler solver steps. 
These proposals are then independently downselected to $K'=6$ representative predictions, following Algorithm~\ref{alg:kmeans}, using KMeans++ initialization~\citep{kmeans2007}. 

\begin{algorithm}[h]
\caption{Weighted K-Means Downselection}
\label{alg:kmeans}
\KwIn{Proposals $\{\hat{\mathbf{Y}}^{(k)}\}_{k=1}^{K}$, target count $K'$, 
      trajectory scale $\alpha=5$}
\KwOut{$K'$ representative predictions}
Extract last-frame features: 
$\mathbf{f}_k = [\alpha \cdot \hat{\mathbf{Y}}^{(k)}_\text{traj}[-1],\; 
\hat{\mathbf{Y}}^{(k)}_\text{pose}[-1]] \in \mathbb{R}^{3 + J \cdot 3}$\;
Initialize $K'$ cluster centers $\{\mathbf{c}_j\}_{j=1}^{K'}$ using 
K-Means++ seeding: select the first center uniformly at random, then 
select each subsequent center with probability proportional to its 
squared distance from the nearest existing center\;
\Repeat{center assignments unchanged \textbf{or} $n_\text{iter} = 30$}{
    Assign each proposal to nearest center: 
    $a_k = \arg\min_j \|\mathbf{f}_k - \mathbf{c}_j\|_2$\;
    Update centers: 
    $\mathbf{c}_j = \frac{1}{|\mathcal{C}_j|}\sum_{k \in \mathcal{C}_j} 
    \mathbf{f}_k$\;
}
\For{each cluster $j$}{
    Select medoid: 
    $k^*_j = \arg\min_{k \in \mathcal{C}_j} \|\mathbf{f}_k - 
    \mathbf{c}_j\|_2$\;
}
\Return $\{\hat{\mathbf{Y}}^{(k^*_j)}\}_{j=1}^{K'}$\;
\end{algorithm}

\paragraph{Evaluation metrics.}
\label{app:metrics}
Let $\hat{\mathbf{Y}}^{(k)} \in \mathbb{R}^{T_\text{out} \times J \times 3}$ denote the $k$-th predicted absolute joint position sequence recovered via B-spline evaluation at the end of ODE integration, as described in \S\ref{sec:spline}. $\mathbf{Y}^* \in \mathbb{R}^{T_\text{out} \times J \times 3}$ is the ground truth sequence, with $K'$ predictions per agent. The root joint is $j=0$. 

\begin{table}[ht]
\centering
\caption{Evaluation metrics. $\hat{\mathbf{Y}}$ denotes the prediction, $T = T_\text{out}$, $j$ indexes joints, $f$ indexes frames.}
\label{tab:metrics}
\small
\begin{tabularx}{\columnwidth}{l l X} 
\toprule
\textbf{Metric} & \textbf{Formula} & \textbf{Notes} \\
\midrule
JPE (mm) & $\min_k \frac{1}{J}\sum_j \|\hat{\mathbf{Y}}^{(k)}_{T,j} - \mathbf{Y}^*_{T,j}\|_2$ & Absolute joint error at final frame \\[10pt]
APE (mm) & $\min_k \frac{1}{J}\sum_j \|\tilde{\mathbf{Y}}^{(k)}_{T,j} - \tilde{\mathbf{Y}}^*_{T,j}\|_2$ & Root-relative; $\tilde{\mathbf{Y}}_{f,j} = \mathbf{Y}_{f,j} - \mathbf{Y}_{f,0}$ \\[10pt]
FDE (mm) & $\min_k \|\hat{\mathbf{Y}}^{(k)}_{T,0} - \mathbf{Y}^*_{T,0}\|_2$ & Root joint displacement at final frame \\[10pt]
NPSS     & $\frac{1}{F}\sum_f \cos\!\left(\frac{P_f(\hat{\mathbf{Y}})}{\|P_f(\hat{\mathbf{Y}})\|},\, \frac{P_f(\mathbf{Y}^*)}{\|P_f(\mathbf{Y}^*)\|}\right)$ & $P_f$: power at freq. $f$ via rFFT~\citep{gopalakrishnan2019neural} \\
\bottomrule
\end{tabularx}
\end{table}

\paragraph{Retraining baselines.}
We detail the retraining procedure for baselines marked with $(^*)$ in the main paper. 
To enable best-of-$K'$ evaluation for the EMPMP model, we extend it to generate $K'=6$ predictions by replicating its embedding and prepending learned mode embeddings before a shared projection head. Each mode thereby receives a uniquely conditioned input, analogous to how T2P combines a shared local encoding with $K'$ attention-derived global embeddings. All other training hyperparameters follow the original paper.

We retrain DuMMF with the original loss weights, modifying the number of learnable ``intent codes'' (diverse modes to predict on) to $K'=6$. We also adjust the diversity loss scale to account for differences in output dimensionality between the original setting and ours. The original scales were set relative to a 52-joint $\times$ 6D $\times$ 25-frame pose representation (7,800 dim.) and a 3D $\times$ 25-frame trajectory (75 dim.). We rescale to yield comparable $\exp(-d/\text{scale})$ gradient magnitudes across our 15-joint skeleton, setting the pose diversity scale to 2,000 and the trajectory diversity scale to 200. 

We retrain T2P on WorldPose after adapting its joint partitioning to the 24-joint SMPL skeleton, reassigning body parts to the nearest anatomically equivalent SMPL joints. We also increase the neighbor search radius used for social context to reflect the larger spatial scale of the data. 

\section{B-Spline Reparameterization Details}
\label{app:bspline}
As introduced in~\S\ref{sec:spline}, rather than predicting all $T_\text{out}$ future frames directly, we reparameterize the output in a compressed control-point space via cubic B-splines, with separate bases for trajectory ($C_t$ control points) and pose ($C_p$ control points).

\paragraph{Clamped knot vector.}
Given $C$ control points and degree $d=3$, we construct a clamped knot vector with $n = C - d - 1$ uniformly spaced interior knots: 
\begin{equation}
  \boldsymbol{\xi} = \bigl[\,\underbrace{0,\ldots,0}_{d+1},\;
  \tfrac{1}{n+1},\;\tfrac{2}{n+1},\;\ldots,\;\tfrac{n}{n+1},\;
  \underbrace{1,\ldots,1}_{d+1}\,\bigr].
\end{equation}
Clamping ensures $\boldsymbol{\xi}$ has $d+1$ repeated knots at each endpoint, guaranteeing the spline interpolates its first and last control points exactly.

\paragraph{Cox-de Boor basis functions.}
The B-spline basis functions $N_{i,k}(\tau)$ are defined recursively:
\begin{align}
  N_{i,0}(\tau) &= \begin{cases} 1 & \text{if } \xi_i \le \tau < \xi_{i+1} \\ 
  0 & \text{otherwise} \end{cases}, \quad\text{and} \\[4pt]
  N_{i,k}(\tau) &= \frac{\tau - \xi_i}{\xi_{i+k} - \xi_i}\,N_{i,k-1}(\tau) 
  \;+\; \frac{\xi_{i+k+1} - \tau}{\xi_{i+k+1} - \xi_{i+1}}\,N_{i+1,k-1}(\tau),
\end{align}
for $k = 1,\ldots,d$, with the convention $0/0 \equiv 0$.

\paragraph{Basis matrix.}
Evaluating all $C$ basis functions at $T_\text{out}$ uniformly spaced parameter values $\tau_m = m/(T_\text{out}-1)$, $m=0,\ldots,T_\text{out}-1$, yields the basis matrix:
\begin{equation}
  \mathbf{B} \in \mathbb{R}^{T_\text{out} \times C}, \qquad \text{with} \quad
  B_{m,i} = N_{i,d}(\tau_m). 
\end{equation}
The forward mapping from control points to full-resolution sequences is $\mathbf{Y} = \mathbf{B}\tilde{\mathbf{Y}}$, and the reverse mapping uses the Moore-Penrose pseudoinverse $\tilde{\mathbf{Y}} = \mathbf{B}^\dagger \mathbf{Y}$, precomputed once before training.

\paragraph{Training projection.}
Ground-truth futures are projected to control-point space before flow matching:
\begin{equation}
  \tilde{\mathbf{Y}}^1_\text{traj} = \mathbf{B}_t^\dagger\,\mathbf{Y}^1_\text{traj}, 
  \qquad
  \tilde{\mathbf{Y}}^1_\text{pose} = \mathbf{B}_p^\dagger\,\mathbf{Y}^1_\text{pose}.
\end{equation}
Noise is sampled directly in control-point space, $\tilde{\mathbf{Y}}^0_\text{traj}, \tilde{\mathbf{Y}}^0_\text{pose} \sim \mathcal{N}(0, \mathbf{I})$,
preserving an isotropic Gaussian source distribution in the compressed 
representation. The interpolant and target velocity are:
\begin{equation}
  \tilde{\mathbf{Y}}^t = (1-t)\,\tilde{\mathbf{Y}}^0 + t\,\tilde{\mathbf{Y}}^1, 
  \qquad\text{and}\quad 
  v^* = \tilde{\mathbf{Y}}^1 - \tilde{\mathbf{Y}}^0.
\end{equation}
The model predicts $\hat{v}^\text{ctrl} = v_\theta(\tilde{\mathbf{Y}}^t, t \mid \mathbf{X})$ 
in control-point space. The velocity residual is projected back to full 
resolution before computing the temporally weighted loss:
\begin{equation}
  \mathbf{e} = \mathbf{B}\,(\hat{v}^\text{ctrl} - v^*) \in \mathbb{R}^{T_\text{out} \times F},
\end{equation}
where $F=3$ for trajectory and $F=J{\times}3$ for pose.

\section{Flow Matching Details.}
\label{app:flow}
\paragraph{Training.} Algorithm~\ref{alg:training} summarises the training procedure. At each step, ground-truth futures are projected to control-point space via the precomputed pseudoinverses $\mathbf{B}_t^\dagger$ and $\mathbf{B}_p^\dagger$, noise is sampled directly in that space, and the model is trained to predict the conditional velocity field. The timestep $t$ is drawn from a logit-normal distribution to concentrate training on intermediate noise levels, where the flow is most uncertain. Velocity residuals are projected back to full resolution via $\mathbf{B}$ before the temporally weighted loss is applied, as described in \S\ref{sec:spline}.

\begin{algorithm}[h]
\caption{Flow Matching Training Step}
\label{alg:training}
\KwIn{Past observations $\mathbf{X}$, ground-truth future 
$(\mathbf{Y}^1_\text{traj}, \mathbf{Y}^1_\text{pose})$}
Project GT to control-point space: 
$\tilde{\mathbf{Y}}^1_\text{traj} = \mathbf{B}_t^\dagger\mathbf{Y}^1_\text{traj}$,
$\tilde{\mathbf{Y}}^1_\text{pose} = \mathbf{B}_p^\dagger\mathbf{Y}^1_\text{pose}$\;
Sample noise in control-point space:
$\tilde{\mathbf{Y}}^0_\text{traj}, \tilde{\mathbf{Y}}^0_\text{pose} \sim \mathcal{N}(0,\mathbf{I})$\;
Sample timestep: $z \sim \mathcal{N}(\mu_t, s_t^2)$, $t = \text{sigmoid}(z)$\;
Interpolate: $\tilde{\mathbf{Y}}^t = (1-t)\tilde{\mathbf{Y}}^0 + 
t\tilde{\mathbf{Y}}^1$\;
Predict velocity: $(\hat{v}^\text{ctrl}_\text{traj}, \hat{v}^\text{ctrl}_\text{pose}) 
= v_\theta(\tilde{\mathbf{Y}}^t, t \mid \mathbf{X})$\;
Project residual to frame space: $\mathbf{e}_\text{traj} = 
\mathbf{B}_t\hat{v}^\text{ctrl}_\text{traj} - v^*_\text{traj}$\;
$\mathbf{e}_\text{pose} = \mathbf{B}_p\hat{v}^\text{ctrl}_\text{pose} - 
v^*_\text{pose}$\;
Compute loss $\mathcal{L}$ via Eq.~\ref{eq:loss} and backpropagate\;
\end{algorithm}

\paragraph{Inference.} Algorithm~\ref{alg:inference} describes single-agent inference. $K$ noise realizations are integrated in parallel using an Euler solver over $S$ steps entirely in control-point space, with full-resolution sequences recovered by a single matrix multiplication at the end. 

\begin{algorithm}[h]
\caption{Flow Matching Inference (Single Agent, $K$ Samples)}
\label{alg:inference}
\KwIn{Past observations $\mathbf{X}$, proposals $K$, steps $S$, 
      select $K'$}
\KwOut{$K'$ diverse full-resolution predictions}
\For{$k = 1, \ldots, K$ \textnormal{(in parallel)}}{
  Sample: $\tilde{\mathbf{Y}}^{0,(k)}_\text{traj}, \tilde{\mathbf{Y}}^{0,(k)}_\text{pose} \sim \mathcal{N}(0,\mathbf{I})$\;
  \For{$s = 0, \ldots, S-1$}{
    $t_s = s/S$,\quad $\Delta t = 1/S$\;
    $(\hat{v}^\text{ctrl}_\text{traj}, \hat{v}^\text{ctrl}_\text{pose}) = 
    v_\theta(\tilde{\mathbf{Y}}^{t_s,(k)}, t_s \mid \mathbf{X})$\;
    $\tilde{\mathbf{Y}}^{t_{s+1},(k)} = \tilde{\mathbf{Y}}^{t_s,(k)} + 
    \hat{v}^\text{ctrl} \cdot \Delta t$\;
  }
  Decode: $\hat{\mathbf{Y}}^{(k)}_\text{traj} = 
  \mathbf{B}_t\tilde{\mathbf{Y}}^{1,(k)}_\text{traj}$,\quad
  $\hat{\mathbf{Y}}^{(k)}_\text{pose} = 
  \mathbf{B}_p\tilde{\mathbf{Y}}^{1,(k)}_\text{pose}$\;
}
Downselect $K \to K'$ via Algorithm~\ref{alg:kmeans}\;
\Return $\{\hat{\mathbf{Y}}^{(k^*_j)}\}_{j=1}^{K'}$\;
\end{algorithm}

\paragraph{Likelihood scoring.}
To score individual predictions for visualization, we estimate the log-likelihood of each proposed prediction under the learned flow using the probability-flow ODE~\citep{lipman2022flow,grathwohl2018ffjord}. Given a prediction $\tilde{\mathbf{Y}}^{1,(k)}$ in control-point space, we integrate the reverse-time ODE
\begin{equation}
    \frac{\mathrm{d}\tilde{\mathbf{Y}}}{\mathrm{d}t} = -v_\theta(\tilde{\mathbf{Y}}, t \mid \mathbf{X})
\end{equation}
from $t=1$ to $t=0$ using $S_\text{ll}=100$ Euler steps to recover the corresponding noise sample $\tilde{\mathbf{Y}}^{0,(k)}$. The change in log-density along the trajectory is accumulated via Hutchinson's trace estimator~\citep{Hutchinson01011990}:
\begin{equation}
    \Delta \log p \approx -\sum_{s} \boldsymbol{\epsilon}^\top 
    \nabla_{\tilde{\mathbf{Y}}} 
    \bigl(\boldsymbol{\epsilon}^\top v_\theta(\tilde{\mathbf{Y}}, t_s \mid \mathbf{X})\bigr)
    \,\Delta t, 
    \qquad \boldsymbol{\epsilon} \sim \{\pm 1\}^d,
\end{equation}
applied jointly to trajectory and pose dimensions. The final log-likelihood is:
\begin{equation}
    \log p(\tilde{\mathbf{Y}}^{1,(k)} \mid \mathbf{X}) =
    \log \mathcal{N}(\tilde{\mathbf{Y}}^{0,(k)}_\text{traj};\,0,\,\mathbf{I})
    + \log \mathcal{N}(\tilde{\mathbf{Y}}^{0,(k)}_\text{pose};\,0,\,\mathbf{I})
    + \Delta \log p.
\end{equation}
This scoring is used only for the visualizations in Figures~\ref{fig:teaser} and~\ref{fig:worldpose_qual}; all evaluation metrics (JPE, APE, FDE, NPSS) use best-of-$K'$ selection without expensive likelihood scoring.

\section{Additional Ablation Studies}
\label{app:ablation}

\begin{figure}[t]
    \centering
    \includegraphics[width=1\linewidth]{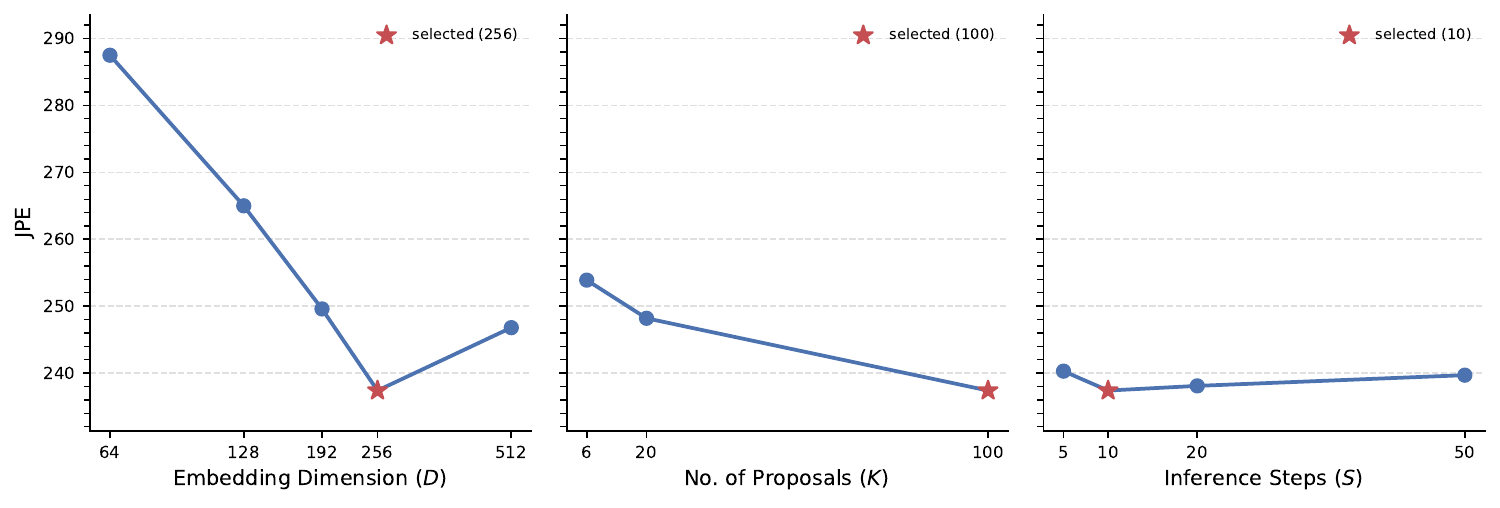}
    \caption{Sensitivity analysis of key hyperparameters on CMU-Mocap (UMPM), 
    reporting JPE at 2.0s (lower is better). Each plot varies one hyperparameter 
    while holding all others fixed at their selected values (marked $\star$). 
    \emph{(Left)} transformer embedding dimension $D$. 
    \emph{(Middle)} number of proposals $K$ for best-of-$K'$ selection. 
    \emph{(Right)} number of Euler integration steps $S$ at inference.}
    \label{fig:add_hp_comp}
\end{figure}
\paragraph{Hyperparameter choices.}
Figure~\ref{fig:add_hp_comp} reports sensitivity to three key hyperparameters. The embedding dimension has the largest impact: underpowered at $D=64$ and beginning to degrade at $D=512$, with $D=256$ offering the best trade-off between accuracy and efficiency. The number of proposals $K$ shows an impact with JPE dropping $\sim16mm$ from $K=6$ to $K=100$, confirming that best-of-$K'$ selection benefits from a large proposal pool.  
Inference steps show near-flat performance across the tested range (5--50 steps), with the minimum at $S=10$, indicating that the ODE integrates reliably with very few steps.

\paragraph{Output-head design.}
On CMU-Mocap at 2.0s, under the same $K'=6$ evaluation protocol as Table~\ref{tab:motion_eval_consolidated}, our full model obtains 237.4\,mm JPE. Replacing independent sampling and downselection with $K'$ direct output heads increases JPE by $\sim18\%$ relative to this baseline. A learned-GMM source distribution with one component per mode slot, an LED-style~\citep{mao2023leapfrog} flow-matching analog of a direct multi-mode initializer, increases JPE by $\sim10\%$. Full-body prediction makes fixed slots restrictive because the trajectory-optimal and pose-optimal modes need not coincide. Our independent trajectory and pose noise draws permit their recombination within the $K=100$ proposal pool, while also allowing $K'$ to change at test time without retraining. Direct heads are cheaper, however, and the latency results below quantify the cost of our sampling-based choice.

\begin{table}[t]
    \centering
    \caption{Total inference latency comparison. Mean $\pm$ std. calculated over 10 runs. All models are run on an NVIDIA L40S GPU with a batch size of 64. Lower values are better $(\downarrow)$. A $(^*)$ denotes models we recomputed for this experiment.}
    \label{tab:latency}
    \small
    \begin{tabularx}{\textwidth}{l >{\centering\arraybackslash}X >{\centering\arraybackslash}X >{\centering\arraybackslash}X}
    \toprule
    \textbf{Model} & 
    \textbf{\makecell{CMU-Mocap (UMPM)\\ Latency (s) $\downarrow$}} & 
    \textbf{\makecell{3DPW\\ Latency (s) $\downarrow$}} & 
    \textbf{\makecell{WorldPose\\ Latency (s) $\downarrow$}} \\
    \midrule
    T2P$^*$   & 8.7 $\pm$ 0.3 & 4.3 $\pm$ 0.9 & 30.1 $\pm$ 2.2\\
    EMPMP$^*$ & 6.5 $\pm$ 0.1 & 3.8 $\pm$ 0.2 & 21.9 $\pm$ 1.0 \\
    \textbf{Ours}    & 304.3 $\pm$ 14.8 & 221.6 $\pm$ 4.7 & 1767.0 $\pm$ 31.3\\
    \bottomrule
    \end{tabularx}
\end{table}
\paragraph{Inference latency.}
Table~\ref{tab:latency} reports total wall-clock inference time for end-to-end evaluation of each test set on a single NVIDIA L40S GPU with batch size 64, including proposal generation, $S=10$ ODE steps, and weighted k-means downselection. CMU-Mocap (UMPM) contains 3,000 test scenes with 3 agents each, 3DPW contains 1,640 scenes with 2 agents, and WorldPose contains 2,252 scenes with up to 22 agents evaluated with $N=10$ neighbor conditioning tokens.

Our method is substantially slower than the deterministic baselines, reflecting the cost of generating $K=100$ proposals per agent via ODE integration before downselection. T2P and EMPMP produce a single prediction per forward pass, whereas our method requires $K$ independent forward passes per agent per integration step, giving an approximate runtime scaling of $K \times S$ relative to a single-pass deterministic method. The WorldPose gap is further amplified by the larger number of agents per scene, higher control-point counts ($C_t=30$, $C_p=45$) required for the longer 3.0s horizon, and additional tokens for $N=10$ neighbors' trajectories.

This latency trade-off is inherent to stochastic prediction with large proposal counts and is common to diffusion and flow matching methods at inference. At the same final $K'=6$ hypothesis budget, however, FlowMoP is $\sim14\times$ faster than DuMMF on CMU-Mocap because its proposals are integrated in parallel. For applications where latency is critical, $K$ can be reduced substantially; our ablation (Fig. ~\ref{fig:add_hp_comp}) shows that reducing $K$ from 100 to 20 incurs only a modest accuracy penalty while cutting runtime proportionally. The per-sample forward pass latency reported in Table~\ref{tab:representation_comparison} (6.6ms) reflects the cost of a single model evaluation, which is competitive with deterministic baselines at matched proposal count.

\paragraph{Collision analysis.}
\label{app:collision}

\begin{wraptable}{r}{0.44\textwidth}
    \centering
    \vspace*{-15pt}
    \caption{Inter-agent collision rate (\%) on WorldPose at different prediction horizons. Lower is better~$(\downarrow)$.}
    \label{tab:collision}
    \small
    \begin{tabularx}{\linewidth}{l @{\extracolsep{\fill}} ccc}
        \toprule
        \textbf{Setting} & \textbf{1.0s} & \textbf{2.0s} & \textbf{3.0s} \\
        \midrule
        Ground truth & 0.14 & 0.13 & 0.13 \\
        Ours $N=0$   & 2.69 & 3.43 & 4.27 \\
        Ours $N=3$   & 2.61 & 3.34 & 4.31 \\
        Ours $N=10$  & 1.64 & 2.09 & 2.65 \\
        \bottomrule
    \end{tabularx}    
    \vspace*{-8pt}
\end{wraptable}

Table~\ref{tab:collision} reports the inter-agent collision rate on WorldPose, defined as any joint pair across 
agents falling within 0.3\,m, measured as the average rate across all $K' \times K'$ prediction pairs per agent pair. GT denotes the collision rate observed in ground-truth future sequences.
The ground truth collision rate ($\sim0.14\%$ across all future horizons) reflects the natural frequency of near-contact events in real soccer play and serves as a reference for physically plausible motion.
Neighbor conditioning substantially reduces collisions in our predictions: $N=10$ achieves a 39\% reduction in collision rate at 1.0s relative to $N=0$, and the gap widens at longer horizons where positional uncertainty accumulates. This suggests that conditioning on observed neighbor trajectories implicitly encourages spatial separation even without an explicit collision penalty, as the model learns that nearby agents tend to move in mutually compatible directions. The remaining gap between $N=10$ and the ground truth rate can be attributed to the lack of global context (i.e., predictions for all 21 other players).

\paragraph{Absolute field location.}
Our agent-centric representation removes absolute field location. Retaining the last observed global offset as an additional conditioning input yields a mixed result at 3.0s on WorldPose: FDE improves from 1.75\,m to 1.71\,m, while APE degrades from 0.10\,m to 0.15\,m. Field position helps locate the root but competes with the root-relative pose objective, motivating our normalized formulation.

\section{Extended Generative Diversity Evaluation}
\label{app:appendix_diversity}

To rigorously assess the stochastic properties and multi-modal coverage of our method, we provide an extended evaluation of its generative diversity compared to DuMMF, the primary stochastic baseline for multi-person 3D motion prediction. 

\paragraph{Average Pairwise Distance (APD).} 
We measure the APD between generated proposals to quantify the absolute spread of the predictive distribution. For each agent, it is computed as the mean pairwise Euclidean distance over joints, averaged across all agent sequences and proposal pairs:
\begin{equation}
    \text{APD} = \frac{1}{\binom{K}{2}} \sum_{i=1}^{K-1} \sum_{j=i+1}^{K} \| \hat{Y}^{(i)} - \hat{Y}^{(j)} \|_2.
\end{equation}

\begin{table}[t]
\centering
\caption{APD in \emph{mm} on CMU-Mocap (UMPM) measured across $K$ samples per agent.}
\label{tab:apd_comparison}
\small
\begin{tabularx}{\textwidth}{X Y Y Y Y Y Y}
\toprule
& \multicolumn{3}{c}{\textbf{$K=10$}} & \multicolumn{3}{c}{\textbf{$K=100$}} \\
\cmidrule(lr){2-4} \cmidrule(lr){5-7}
\textbf{Model} & \textbf{<1.0s} & \textbf{1.0-2.0s} & \textbf{Total} & \textbf{<1.0s} & \textbf{1.0-2.0s} & \textbf{Total} \\
\midrule
DuMMF & $110.3$ & $312.4$ & $211.3$ & $121.3$ & $314.2$ & $222.8$ \\
Ours & 129.8 & $242.0$ & $185.9$ & 140.7 & $256.9$ & $198.4$ \\
\bottomrule
\end{tabularx}
\end{table}

Table~\ref{tab:apd_comparison} reports APD across $K \in \{10, 100\}$ proposals per agent, averaged over the first-second window (frames 0-25), the second-second window (frames 25-50), and the full 2.0s sequence. Proposals are independent noise draws for our model; for DuMMF, $K=100$ is obtained via multiple forward passes of $K'=6$ learned ``intent codes'' and is therefore not fully independent. 
While DuMMF exhibits higher absolute APD at the final 2.0s horizon (314.2 vs.\ 256.9 at $K=100$), our method yields higher diversity at the 1.0s mark (140.7 vs.\ 121.3). This indicates that while DuMMF's autoregressive rollout produces a wider spread of poses at the end of the temporal horizon, possibly symptomatic of cumulative temporal drift in the autoregressive prediction.

\paragraph{Sample Efficiency and Mode Coverage.}
\begin{figure}[t]
    \centering
    \includegraphics[width=\textwidth]{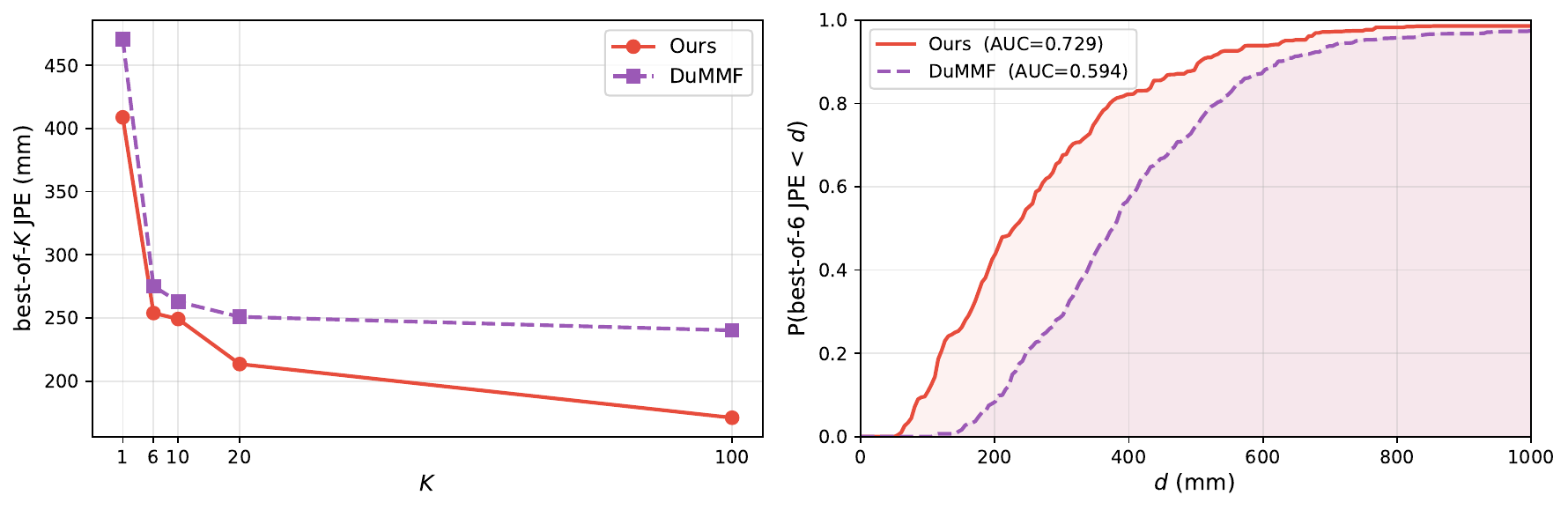}
    \caption{Generative diversity evaluation on CMU-Mocap (UMPM). \emph{(Left)} Best-of-$K$ JPE versus proposal budget $K$ (lower is better). \emph{(Right)} Coverage recall CDF. The plot displays $P(\min_{k'} \text{JPE} < d)$ for $k'\in K'$ predictions, plotted as a function of threshold $d$. Higher and further left is better.}
    \label{fig:generative_diversity}
\end{figure}

To demonstrate that our model's diversity represents contextually grounded multi-modality rather than arbitrary variance, we evaluate sample efficiency and mode coverage (Figure~\ref{fig:generative_diversity}). 
Our method consistently outperforms DuMMF across $K$ on best-of-$K$ JPE. Notably, while DuMMF exhibits a sharp initial reduction in error up to $K=6$, its performance curve subsequently flattens. This plateau is likely an artifact of their training objective, which explicitly optimizes for a predefined number of diverse samples, in this case, 6. In contrast, our continuous flow matching formulation does not impose such a bottleneck on mode count; at $K=10$, our method improves on DuMMF's JPE at $K=20$ (250.9mm).

We also study the Coverage Recall CDF to disentangle useful diversity from unbounded variance. Evaluated at $K'=6$ diverse representative predictions, this metric measures the empirical probability that the minimum-error prediction falls within a distance $d$ of the ground truth. Our method dominates DuMMF, yielding a significantly higher area under the curve (AUC) of $0.729$ compared to $0.594$. This confirms that our framework generates a rich, multi-modal distribution that accurately covers the true data modes.

\begin{table}[t]
\centering
\caption{Distributional quality and calibration on CMU-Mocap (UMPM) using $K'=6$ samples. Lower is better~$(\downarrow)$ for energy score and CRPS. A spread/error ratio of 0.5 is ideal.}
\label{tab:proper_scores}
\small
\begin{tabular}{l ccc}
\toprule
\textbf{Model} & \textbf{Energy score (mm) $\downarrow$} & \textbf{CRPS (mm) $\downarrow$} & \textbf{Spread/error} \\
\midrule
DuMMF & 5491 & 96.2 & 0.62 \\
\textbf{Ours} & \textbf{4218} & \textbf{61.4} & \textbf{0.48} \\
\bottomrule
\end{tabular}
\end{table}

\paragraph{Energy scoring and calibration.}
To assess how our predictions' probability mass is distributed, we compare our energy score and per-coordinate CRPS over the full 2.0s CMU-Mocap sequence with DuMMF with $K'=6$ sample budget:
\begin{equation}
  \mathrm{ES}=\frac{1}{K'}\sum_{k=1}^{K'}\|\hat{\mathbf{Y}}^{(k)}-\mathbf{Y}^*\|_2
  -\frac{1}{2K'^2}\sum_{k=1}^{K'}\sum_{\ell=1}^{K'}
  \|\hat{\mathbf{Y}}^{(k)}-\hat{\mathbf{Y}}^{(\ell)}\|_2,
\end{equation}

Table~\ref{tab:proper_scores} shows that our model has a 23\% lower energy score and substantially lower coordinate-wise CRPS than DuMMF. We also compare predictive spread as a scalar reliability diagnostic, with a mean error of 0.5 indicating matched dispersion, while values above 0.5 indicate over-dispersion. Our ratio of 0.48 is close to ideal, whereas DuMMF's 0.62 is over-dispersed. Together with the coverage CDF, this shows that our distribution is more diverse, better placed, and better scaled relative to the observed future.

\section{Additional Qualitative Results}
\label{app:qual}
\begin{figure}[t]
    \centering
    \includegraphics[width=\linewidth]{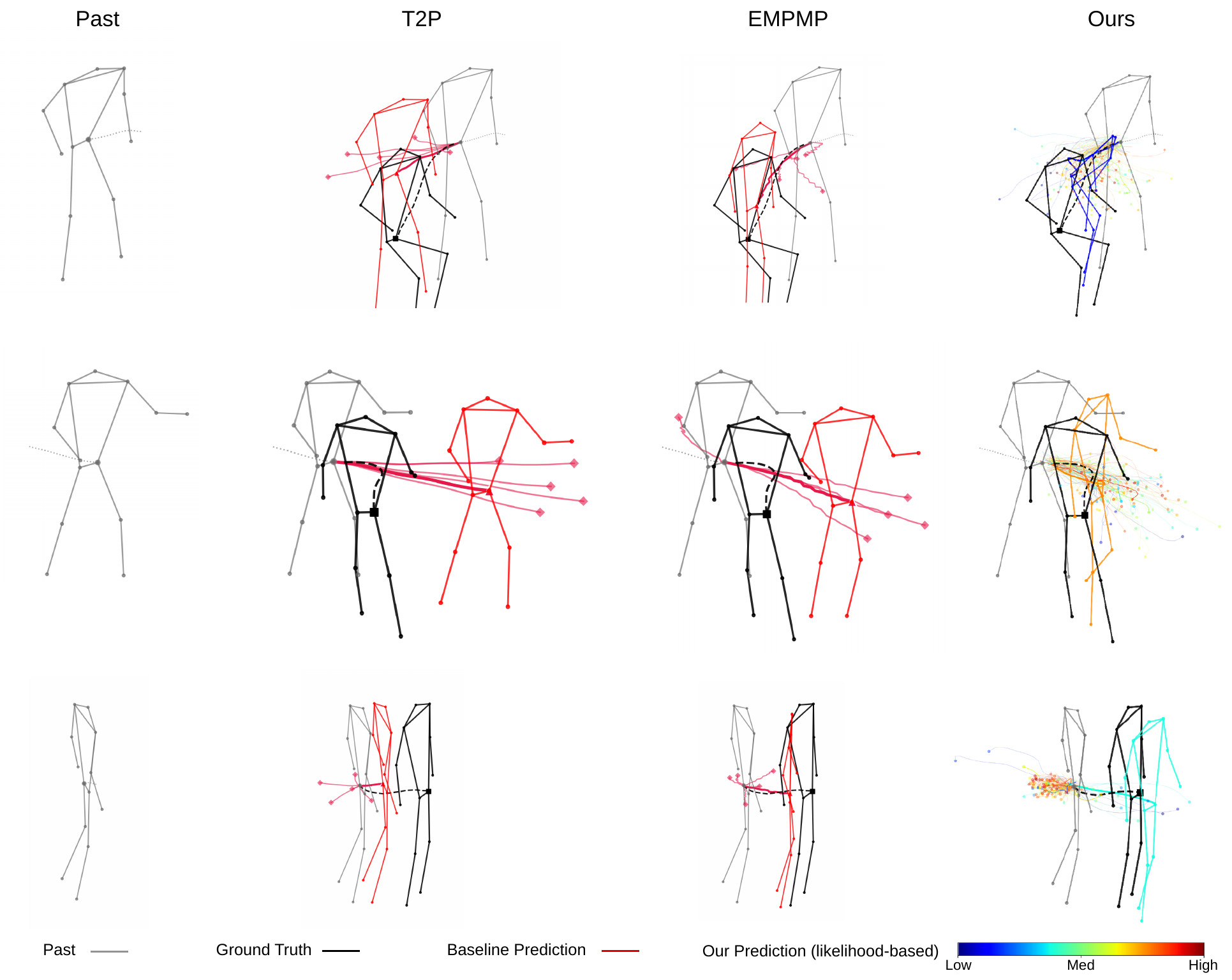}
    \caption{Qualitative comparison on 3DPW. Each column shows the root trajectory and final-frame skeleton for a single test agent: the observed past, baselines, and our model. Dashed black lines denote ground truth future trajectory; colored lines show the trajectories of predictions.}
    \label{fig:3dpw_qual}
\end{figure}

Figure~\ref{fig:3dpw_qual} shows qualitative predictions on 3DPW across three test agents exhibiting unconventional motion: a sitting transition, an unexpected change in trajectory direction, and a stationary pose with torso rotation. These cases highlight a key advantage of the generative formulation; the model assigns non-negligible probability mass to low-likelihood but plausible continuations, capturing motion modes that deterministic methods cannot represent. T2P collapses to near-static translated poses, as noted in the main paper, while EMPMP produces more accurate single predictions but loses detail on physical constraints.

\begin{figure}[t]
    \centering
    \includegraphics[width=\linewidth]{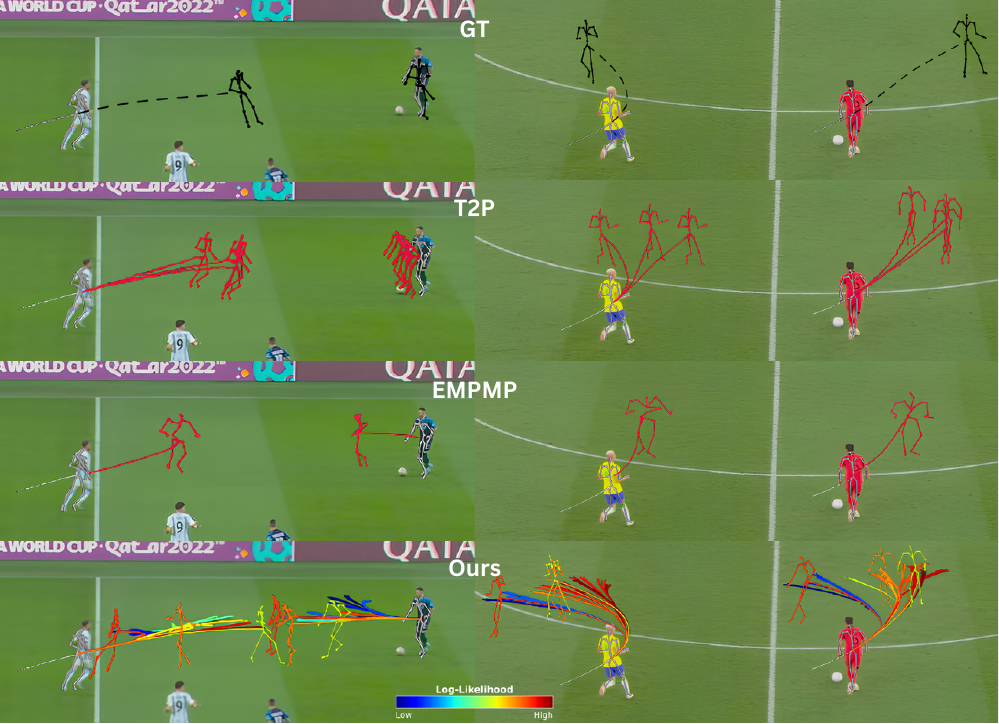}
    \caption{Qualitative comparison on WorldPose. The gray skeleton and trails mark the recent motion history. Each row shows the root trajectory and final-frame skeleton prediction for a single test agent: the ground truth (black), baselines (red), and our model (red: high, blue: low).}
    \label{fig:worldpose_qual}
\end{figure}

Figure~\ref{fig:worldpose_qual} shows two soccer scenes from WorldPose. To focus the comparison on prediction quality rather than failure modes, we filter each baseline's predictions by final-frame pose error, retaining only those within a reasonable threshold; this yields 3 out of 6 predictions for T2P and 1 out of 6 for EMPMP, reflecting the higher rate of degenerate outputs in the more challenging sports setting. Our model produces a distribution of plausible futures for each agent, which is particularly valuable in a sports context: tactical decision-making and downstream planning benefit from knowing not just the most likely trajectory but the full range of plausible player movements and their relative likelihoods.

\section{Limitations}
\label{app:limitation}

The primary practical constraint of our approach is inference latency: generating $K=100$ proposals per agent via ODE integration is substantially slower than deterministic single-pass methods, as detailed in Appendix Table~\ref{tab:latency}. While $K$ and $S$ can be reduced to trade accuracy for speed, the fundamental cost of stochastic sampling remains higher than deterministic prediction.

Our evaluation is limited to three datasets, CMU-Mocap (UMPM), 3DPW, and WorldPose, covering laboratory, in-the-wild, and broadcast video settings, respectively. Generalization to other motion domains, such as dance, martial arts, or team sports with different interaction structures, remains untested.

Our interaction conditioning relies on observed neighbor trajectories, which provide strong cues for positional coordination but do not capture object-mediated interactions. In ball-sport settings, player motion is often governed by the ball's position and trajectory rather than by direct inter-agent proximity alone. While our architecture supports additional modality tokens and could naturally incorporate ball trajectories as an extra conditioning stream, the WorldPose dataset does not provide ball position annotations, precluding this extension in our current evaluation.

Our fixed cubic B-spline basis may smooth abrupt events such as impacts or hard cuts when too few control points are used. Adaptive control-point counts or non-uniform knots could better preserve such high-curvature motion. We also have not evaluated horizons beyond 3 seconds; longer forecasts may require more control points or autoregressive extension, whose error accumulation remains to be studied.

\begin{figure}[t]
    \centering
    \includegraphics[width=\linewidth]{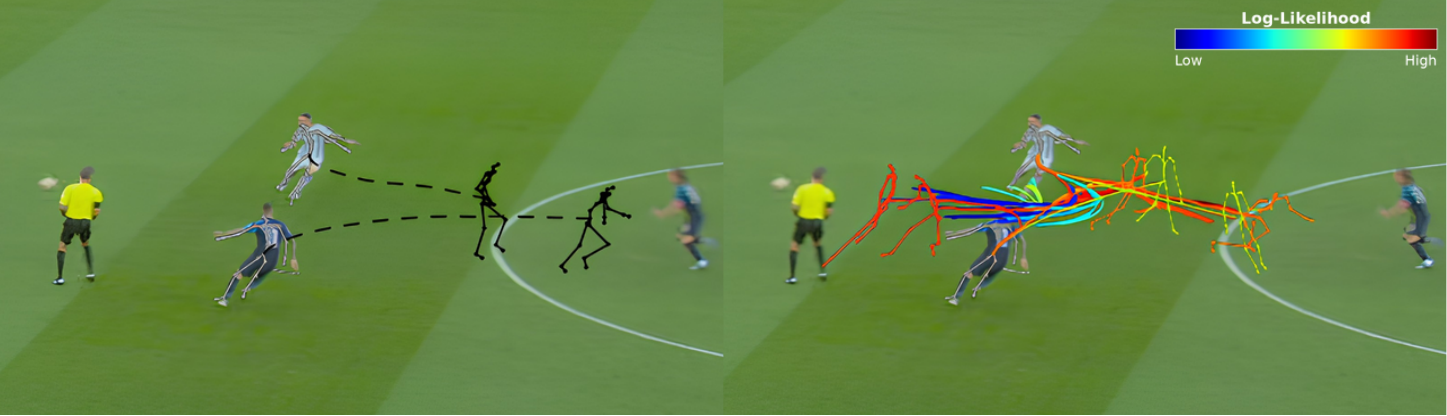}
    \caption{Failure case on WorldPose. Two players (gray trails: observed history) change direction to receive an incoming ball. 
    Ground-truth future paths are shown as dashed black lines, and our model's predicted trajectories are colored by likelihood.}
    \label{fig:failure}
\end{figure}

Finally, our per-agent formulation does not enforce mutual consistency across simultaneously predicted agent futures. In scenarios requiring globally coherent scene predictions, such as collision avoidance or coordinated team plays, this limitation may become significant, and efficient joint prediction over all agents remains an important direction for future work. Additionally, the model imposes no explicit physical constraints, and predicted poses may occasionally exhibit artifacts such as foot skating or self-interpenetration; incorporating physics-based supervision or contact modeling remains an open direction.
Figure~\ref{fig:failure} illustrates a representative failure case: two players alter their paths to receive an incoming ball, but our model predicts several paths that intersect or converge to the same spatial location without awareness of the other agent's predicted motion.

\section{Broader Impacts}
\label{app:impacts}

This work advances stochastic multi-person motion prediction, with potential benefits across several application domains. In sports analytics, calibrated predictive distributions over player motion can support tactical analysis and coaching. In robotics and autonomous navigation, accurate multi-person motion forecasting improves the safety and social compliance of robots operating in shared spaces. In assistive technology, predicting the motion of nearby people can help mobility aids anticipate and avoid 
collisions.

The primary negative societal impact is the potential misuse of motion prediction for surveillance and tracking of individuals without their consent. Our method operates on skeletal pose sequences rather than raw video, which reduces identifiability compared to appearance-based methods, but the underlying capability to track and predict human movement could still be applied in privacy-violating ways. We encourage responsible deployment and adherence to applicable data protection regulations.

\section{Asset Licenses}
\label{app:licenses}

CMU-Mocap~\citep{cmu2003mocap} is freely available for research and non-commercial use. UMPM~\citep{vanderaa2011umpm} is available for non-commercial research purposes. 3DPW~\citep{vonmarcard2018recovering} is available for non-commercial research use. WorldPose~\citep{jiang2024worldpose} is available upon request from the original authors for research purposes.

\end{document}